\documentclass[12pt]{article}
\usepackage{amsmath}
\usepackage{graphicx}
\usepackage{natbib}
\usepackage{url} 

\newcommand{\blind}{0}

\usepackage{epsfig}

\usepackage[utf8]{inputenc}
\usepackage{geometry}
\usepackage{amsmath}\usepackage{amsfonts}\usepackage{amssymb}\usepackage{mathrsfs}
\usepackage{amsthm}
\usepackage{natbib}
\usepackage{listings}\usepackage{color}

\usepackage{epigraph}

\usepackage{newtxmath}   
\usepackage{float}
\usepackage[framemethod=TikZ]{mdframed}

\usetikzlibrary{fit,positioning,arrows.meta}
\usepackage{tikz}
\usetikzlibrary{arrows,shapes,backgrounds}
\usetikzlibrary{decorations.markings}
\usetikzlibrary{calc}
\usetikzlibrary{positioning}
\usetikzlibrary{math}
\usepackage{xcolor}
\usepackage{hyperref}
\usepackage{multimedia}
\usepackage{comment}
\usepackage{bbm}
\usepackage{bm}
\usepackage[english]{babel}
\usepackage{blindtext}
\usepackage{microtype}

\usepackage{caption}
\usepackage{subcaption}

\newcommand{\beq}{\begin{equation}}
\newcommand{\eeq}{\end{equation}}
\newcommand{\bea}{\begin{eqnarray}}
\newcommand{\eea}{\end{eqnarray}}

\newcommand{\be}{\begin{equation}}
\newcommand{\ee}{\end{equation}}

\newmdenv[%
    backgroundcolor=red!8,
    linecolor=red,
    outerlinewidth=1pt,
    roundcorner=5mm,
    skipabove=\baselineskip,
    skipbelow=\baselineskip,
]{redbox}

\newcommand{\x}{\bm{x}}
\newcommand{\y}{\bm{y}}

\newcommand{\h}{\bm{h}}
\newcommand{\bias}{\bm{b}}
\newcommand{\W}{\bm{W}}
\newcommand{\U}{\bm{U}}

\newcommand{\given}{\ | \ }
\newcommand{\act}{g} 

\newtheorem{thm}{Theorem}
\newtheorem{defn}{Definition}

\newtheorem{remark}{Remark}

\begin{document}
\def\spacingset#1{\renewcommand{\baselinestretch}%
{#1}\small\normalsize} \spacingset{1}

\if0\blind
{
\title{\bf Industrial Forecasting with Exponentially Smoothed Recurrent Neural Networks}
  
  \author{Matthew Dixon\thanks{Department of Applied Mathematics, Illinois Institute of Technology, Chicago; matthew.dixon@iit.edu.}}
  \maketitle
} \fi
\if1\blind
{
  \bigskip
  \bigskip
  \bigskip
  \begin{center}
    {\LARGE\bf Time Series Modeling with Exponentially Smoothed Recurrent Neural Networks}
\end{center}
  \medskip
} \fi

\bigskip

\begin{abstract}

Time series modeling has entered an era of unprecedented growth in the size and complexity of data which require new modeling approaches. While many new general purpose machine learning approaches have emerged \citep{bayer,Graves2013, pascanu2012difficulty}, they remain poorly understand and irreconcilable with more traditional statistical modeling approaches \citep{boxjen76, GVK126800421}.  We present a general class of exponential smoothed recurrent neural networks (RNNs) which are well suited to modeling non-stationary dynamical systems arising in industrial applications. In particular, we analyze their capacity to characterize the non-linear partial autocorrelation structure of time series and directly capture dynamic effects such as seasonality and trends.  Application of exponentially smoothed RNNs to forecasting electricity load, weather data, and stock prices highlight the efficacy of exponential smoothing of the hidden state for multi-step time series forecasting. The results also suggest that popular, but more complicated neural network architectures originally designed for speech processing, such as LSTMs and GRUs, are likely over-engineered for industrial forecasting and light-weight exponentially smoothed architectures, trained in a fraction of the time, capture the salient features while being superior and more robust than simple RNNs and ARIMA models. Additionally uncertainty quantification of the exponential smoothed recurrent neural networks, provided by Bayesian estimation, is shown to provide improved coverage.
\end{abstract}


\noindent%
\vfill

\newpage
\spacingset{2} 

\section{Introduction}
Recurrent neural networks (RNNs) are the building blocks of modern sequential learning.
RNNs use recurrent layers to capture non-linear temporal dependencies with a relatively
small number of parameters \citep{Graves2013}. They learn temporal dynamics by mapping an input sequence
to a hidden state sequence and outputs, via a recurrent layer and a feedforward layer.  However, despite the success of these and their successors, there appears to be a chasm between the statistical modeling literature (see e.g. \cite{boxjen76, UBMA_276336909, GVK126800421}) and the machine learning literature (see e.g. \cite{bayer, pascanu2012difficulty,Hochreiter1997, schmidhuber1997long}). 

While there have been many recent theoretical developments in recurrent neural networks from a dynamical systems perspective \citep{lim2019predicting, chen2019symplectic,niu2019recurrent}, there are still open fundamental questions as to how the type and properties of the time series data informs the choice of architectures. In particular the merit of combining exponential smoothing with recurrent architectures is a promising direction, but poorly understood.  \cite{mcdermott2018deep} find empirical evidence that Echo State Networks (ESNs) are well suited for spatio-temporal modeling and Uber \citep{SMYL202075} won the 2019 M4 forecasting competition with a hybrid exponential smoothing-RNN method, which exponentiates the output of a neural network with an optimized smoothing parameter and combines it with a past observed time series level. The choice to smooth the output however does not guarantee that the network will be stable and the smoothing parameter is fixed over time, suggesting that it may be limited for use on non-stationary data.

Leaky integrator neurons are continuous dynamical systems which generalize ESNs with a leaking rate parameter which reduces their excitation and forces the reservoir units to change slowly \citep{jaeger2007}. The leaking rate parameter can be chosen to behave like exponential smoothing, although the parameter is not optimized during weight fitting. However, ESNs are designed for short-term memory (STM) capacity and have difficulties with varying time series lengths in the input sequence. Additionally renormalization of states is needed as an additional step to avoid the network becoming oscillatory, and only a subset of the weights can be fitted if the network is to be stable.

Aside from some of the aforementioned limitations of these methods, there are still open fundamental questions as to how the type and properties of the time series data inform the choice of these architectures. One of the main contributions of this paper is to introduce a new class of exponentially smoothed RNNs, with supporting theoretical justification for architectural design choices. This new RNN class, referred to as $\alpha$-RNNs, differs from leaky integrator units in a number of important ways. First, all the weights in $\alpha$-RNNs are trainable, not just the output weights. Second, ESNs have feedback weights from the output to the inputs that do not exist in an $\alpha$-RNN model. Third, the  smoothing parameter, optimized during network training, is applied to the hidden state and can be shown to be necessary for long-term dependencies. Moreover, it is extended to be adaptive and therefore suitable for non-stationary data.

Finally, leaky integration units include noise terms and must therefore be sampled to produce forecasts. On the other hand, $\alpha$-RNNs are deterministic. We also note that the third point is a distinguishing feature over the hybrid exponential smoothing method of  \citep{SMYL202075} which is fixed and limited to stationary data when combined with simple RNNs.


A second challenge with applying neural networks to time series data is the over-reliance on extensive hyper-parameter tuning, a computationally complex global optimization problem with undesirable outcomes which are dependent on initial conditions for parameter choices. Moreover, little is known about how often to re-tune the parameters and the length of historical data used to train the model can affect the hyper-parameter tuning. 

A third challenge is how to combine more traditional and informative diagnostics of time series data, such as stationarity and memory cut-off, into the model selection procedure. Such an approach is standard in the time series modeling literature \cite{boxjen76} but absent in the machine learning literature. Most of the aforementioned studies are engineering orientated and do not provide substantive insight to justify why the architectural choice is well suited to the dataset. More informative diagnostics on RNN fitting are also needed as there is relatively little literature on how to rank the importance of the lagged inputs. There have, however, been a number of recent advances in statistical interpretability for feedforward architectures such as the work of \cite{horel2020significance} who develop tests for the statistical significance of the input variables. Moreover several studies use the analytic gradients of the networks w.r.t. the inputs to rank normalized input variables and there is no reason, in principle, why this can't be extended to RNNs \citep{doi:10.1137/20M1330518}. We also note that there are other techniques in the computer science literature for ranking the importance of lagged inputs for complex recurrent architectures such as LSTMs \citep{DBLP:journals/corr/abs-1905-12034}.

One of the main contributions of this paper is to cast RNNs into a time series modeling framework and rely on statistical diagnostics in combination with cross-validation to tune the architecture. The statistical tests characterize stationarity and memory cut-off length and provide insight into whether the data is suitable for longer-term forecasting and whether the model must be non-stationarity.

It is well known that plain RNNs have difficultly in learning long-term dynamics, due in part to the vanishing and exploding gradients that can result from back propagating the gradients down through the many unfolded layers of the network \citep{bayer, pascanu2012difficulty}. A particular type of RNN, called a Long Short-Term Memory (LSTM) \citep{Hochreiter1997, schmidhuber1997long} was proposed to address this issue of vanishing or exploding gradients which essentially arises due to the shape of the activation function. A memory unit used in a LSTM allows the network to learn which previous states can be forgotten and alleviates the vanishing gradient.  Partly for this reason, LSTMs have demonstrated much empirical success in the literature \citep{Gers2001,7926112}.

The inclusion of forget, reset and update gates in the LSTM, and a slightly simpler variant --- Gated Recurrent Units (GRUs) \citep{DBLP:journals/corr/ChungGCB14}, provides a switching mechanism to forget memory while simultaneously updating a hidden state vector. These units do not, however, provide an easily accessible mathematical structure from which to study their time series modeling properties.  As such, there is much opaqueness about the types of architectures which are best suited to prediction tasks based on data properties such as wide sense non-stationarity (see supplementary material). However we shall show that exponential smoothing not only alleviates the gradient problem but characterizes the time series modeling properties of these architectures.

The main outcome of applying this approach is that it partially identifies the choice of architecture based on both its time series properties and that of the data. The approach is general and easily extensible to GRUs and LSTMs with the inclusion of the reset gate and cell memory. In this paper, the input sequence is assumed to be of fixed length, although the methodology could be extended to variable length as in sequence to sequence learning models \citep{NIPS2014_5346}. 

Finally, an important aspect of forecasting, not addressed above, is uncertainty quantification. When cast in a Bayesian framework, it is well known that deep networks can provide aleatoric uncertainty quantification \citep{polson2017}. However, there is comparatively little statistics literature on the reliability of Bayesian recurrent networks for uncertainty quantification beyond the study of ESNs in \citep{mcdermott2018deep}. 

A summary of the main contributions of this paper are

\begin{itemize}
    \item We show how plain RNNs, with short-term memory, can be generalized to exhibit long-term memory with a smoothing scalar $\alpha$---smoothing also helps offset the infamous vanishing gradient problem in plain RNNs with only one extra parameter;
    \item We show how time series analysis guides the architectural parameters ---the sequence length of the RNN can be determined from the estimated partial autocorrelogram, and tanh activation of the recurrent layer is needed for stability; 
    \item We demonstrate how a dynamic $\alpha_t$-RNN model for non-stationary data, a lightweight version of GRUs and LSTMs, has the ability to model complex non-stationary times series data with comparable performance; and
    \item Finally we assess the performance of Bayesian recurrent networks by estimating the coverage of the confidence intervals at various predictive horizons during back-testing.
\end{itemize}

The remainder of this paper is outlined as follows. Section \ref{sect:rnn} introduces the $\alpha$-RNN and Section \ref{sect:tsm_intro} applies time series analysis to guide the architectural properties. Section \ref{sect:alpha_t} introduces a dynamic version of the model and illustrates the dynamical behavior of $\alpha$. Section \ref{sect:bayesian} describes the methodology for training Bayesian networks and using them for prediction and uncertainty quantification. Details of the training, implementation and experiments together with the results are presented in Section \ref{sect:results}. Finally, Section \ref{sect:conclusion} concludes with directions for future research.

\section{alpha-RNNs} \label{sect:rnn}

\begin{figure}[H]
\begin{subfigure}[t]{0.31\columnwidth}
\centering
\includegraphics[width=\textwidth, height= 0.2\textheight]{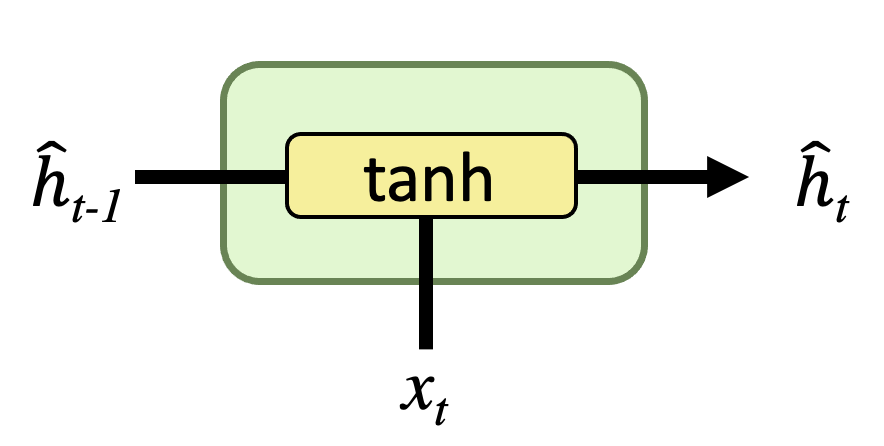} 
\caption{RNN cell.}
\end{subfigure}
\begin{subfigure}[t]{0.31\columnwidth}
\centering
\includegraphics[width=\textwidth, height= 0.2\textheight]{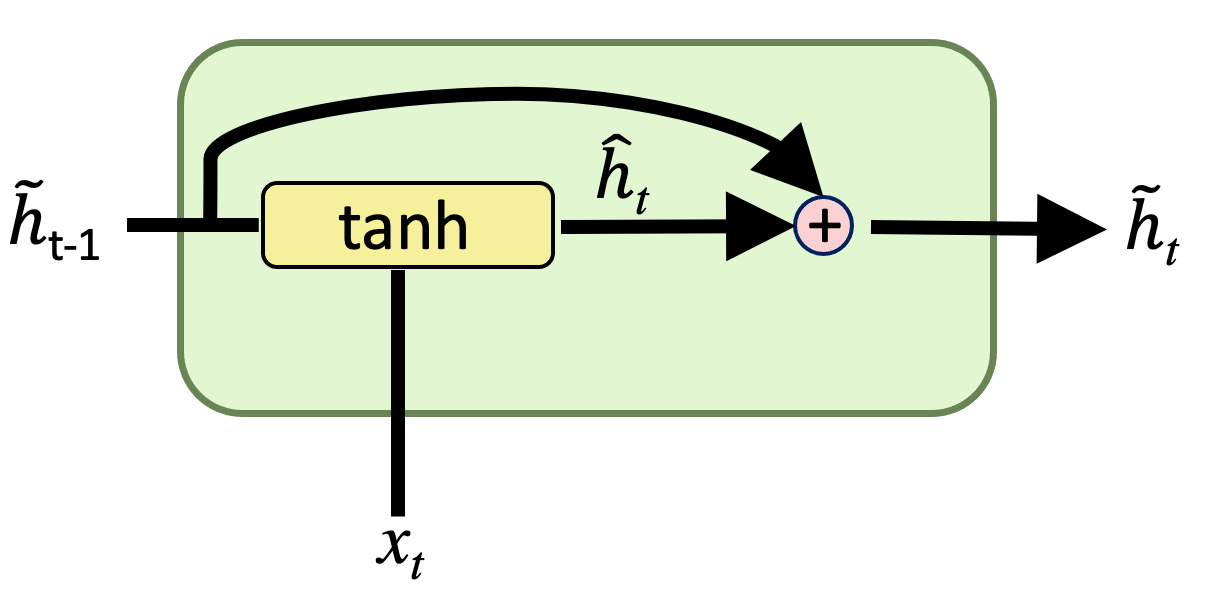} 
\caption{$\alpha$-RNN cell.}
\end{subfigure}
\begin{subfigure}[t]{0.31\columnwidth}
\centering
\includegraphics[width=\textwidth, height= 0.2\textheight]{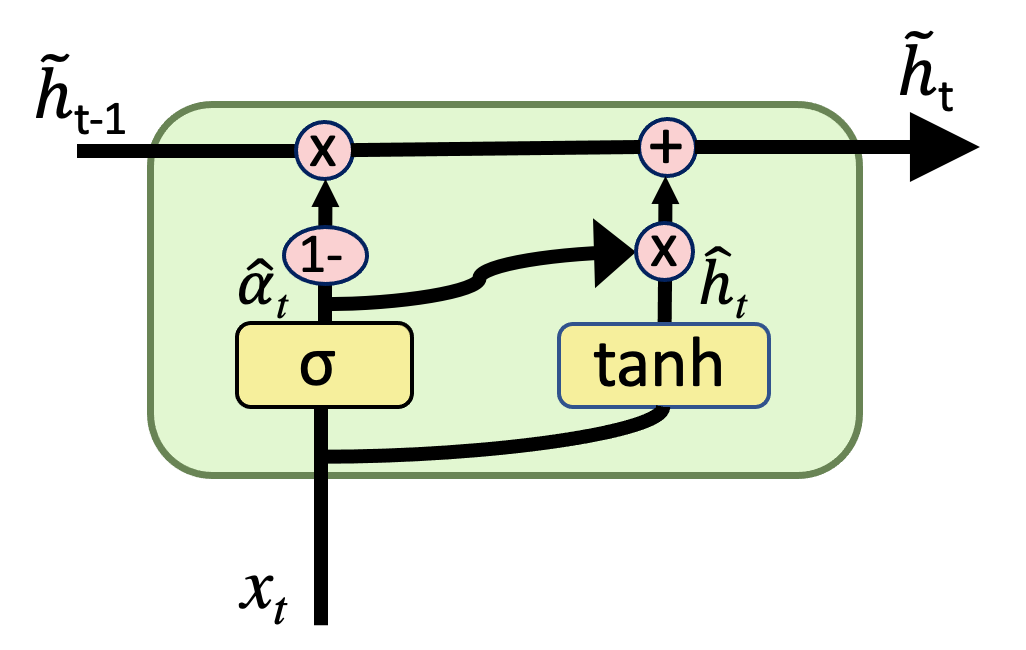} 
\caption{$\alpha_t$-RNN cell.}
\end{subfigure}
\newline
\begin{subfigure}[t]{0.49\columnwidth}
\centering
\includegraphics[width=0.9\textwidth, height= 0.25\textheight]{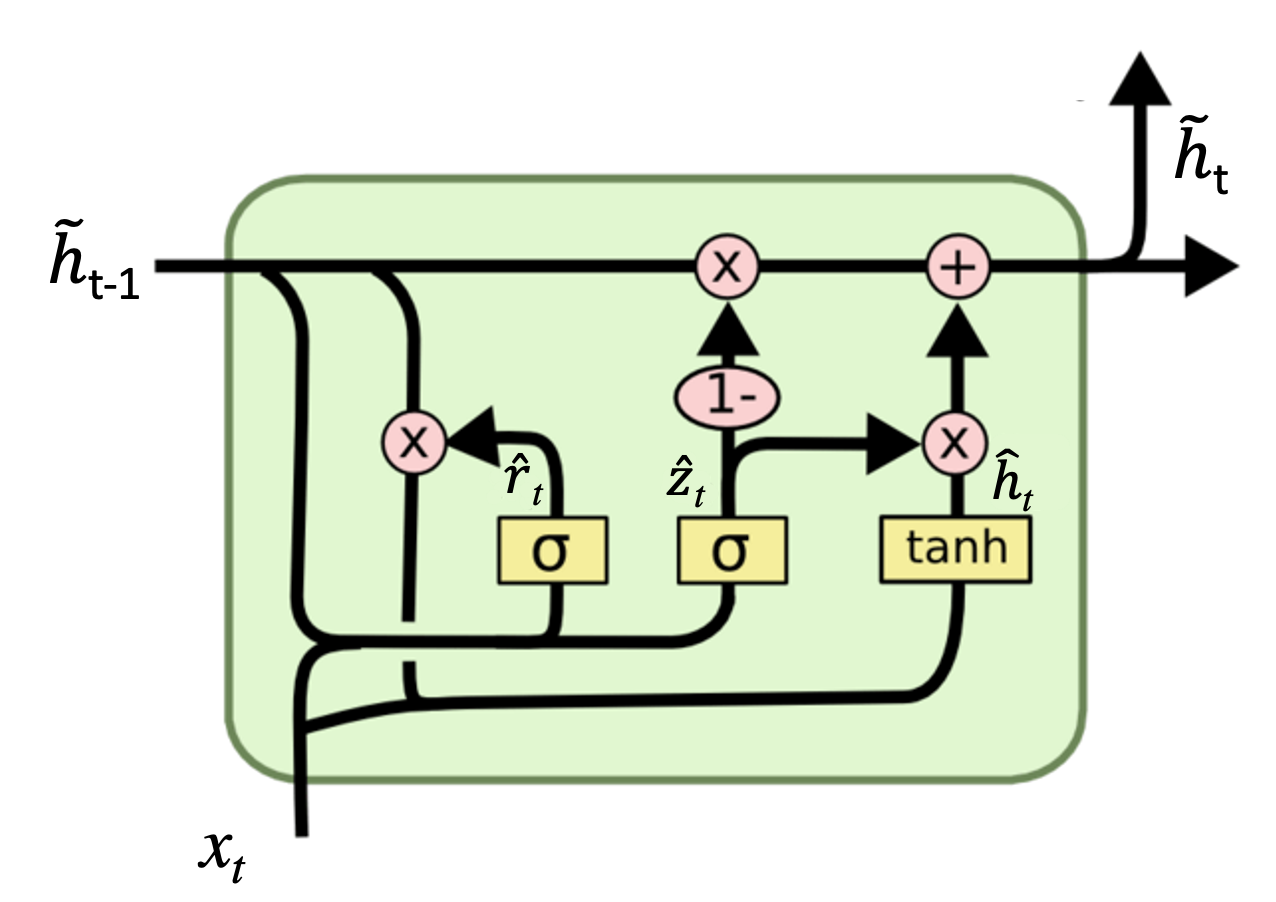} 
\caption{GRU cell.}
\end{subfigure}
\begin{subfigure}[t]{0.49\columnwidth}
\centering
\includegraphics[width=\textwidth, height= 0.25\textheight]{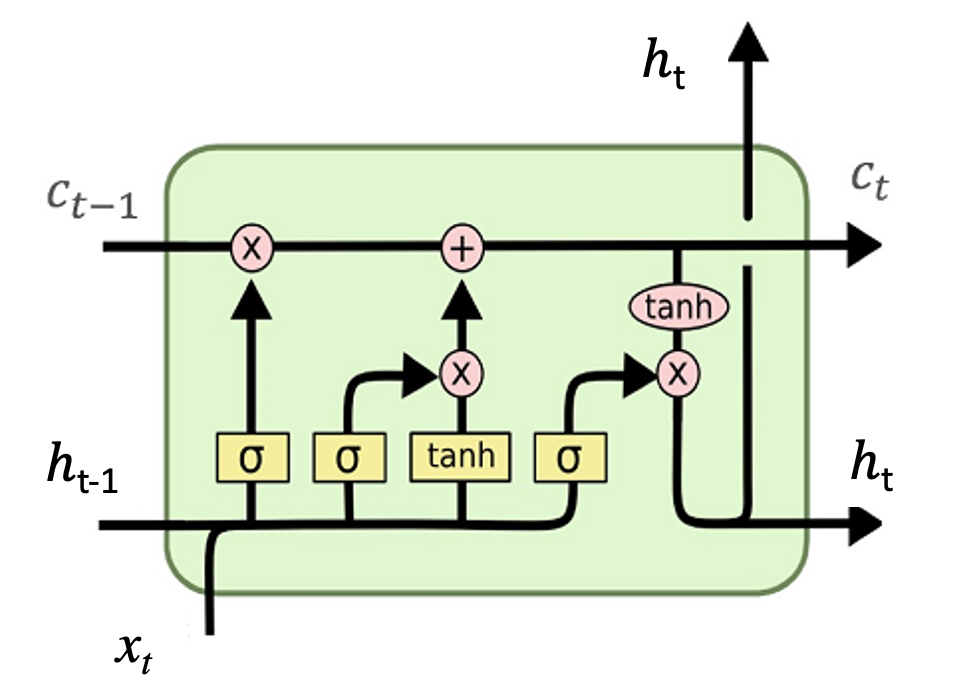} 
\caption{LSTM cell.}
\end{subfigure}

\caption{\textit{A comparison of various recurrent cells starting with the simplest through to the most complex. Arithmetic operations are applied element wise and "$1-$" denotes one minus the input. $\alpha_t$-RNNs are dynamically smoothed $\alpha$-RNNs and are described in Section 4.1. See Appendix D for a description of GRUs and LSTMs. Images in (d) and (e) are slightly modified and credited to Wikimedia Commons.}}
\label{fig:rnn}
\end{figure}

Given auto-correlated observations of covariates or predictors, $\bm{x}_t$, and continuous responses $\bm{y}_t$ at times $t=1,\ldots, N$,  in the time series data 
$\mathcal{D} := \{(\x_t,\y_t) \}_{t=1}^N$,  our goal is to construct  an m-step ($m>0$) ahead times series predictor,$\hat{\y}_{t+m}=F(\bm{\underbar{x}}_t)$, of an observed target, 
$\y_{t+m}\in\mathbb{R}^n$, from a $p$ length input sequence $\bm{\underbar{x}}_t $ 
$$
\y_{t+m} = F(\bm{\underbar{x}}_t) + \mathbf{\epsilon}_t  \qquad {\rm where} \qquad \bm{\underbar{x}}_t:=\{\x_{t-p+1}, \ldots, \x_t \},
$$
$L^j[\x_t]:=\x_{t-j}$ is the $j^{th}$ lagged observation of $\textbf{x}_t\in \mathbb{R}^d$, for $j=0,\ldots, {p-1}$ and $\mathbf{\epsilon}_t\in\mathbb{R}^n$ is the homoscedastic model error at time $t$. We introduce the $\alpha$-RNN model (as shown in Figure \ref{fig:rnn}b):
\begin{eqnarray}
	\hat{\y}_{t+m} &=&F_{W, b, \alpha}(\bm{\underbar{x}}_t),  
\end{eqnarray}
where $F_{W, b, \alpha}(\bm{\underbar{x}}_t)$ is an $\alpha\in[0,1]$ smoothed RNN with weight matrices $\W:=(\W_h, \U_h, \W_y)$ - the input weight matrix $W_h\in \mathbb{R}^{H\times d}$, the recurrence weight matrix $\U_h\in\mathbb{R}^{H\times H}$, the output weight matrix $\W_y\in \mathbb{R}^{n\times H}$, and $H$ is the number of hidden units. The hidden and output bias vectors in $\mathbb{R}^H$ and $\mathbb{R}^n$ are respectively given by biases vectors $\bias:=(\bias_h,\bias_y)$.

For each time step in the sequence $s=t-p+2,\dots,t$, forward passes separately update a hidden \emph{internal} state $\hat{\h}_s \in \mathbb{R}^H$, using the recurrence relations:
\begin{eqnarray*}
\hat{\h}_s &=& \act(\W_h\x_s+ \U_h \tilde{\h}_{s-1} + \bias_h),\\
\tilde{\h}_{s} &=& \alpha \hat{\h}_{s} + (1-\alpha)\tilde{\h}_{s-1},
\end{eqnarray*}
where $\act(\cdot)$ is a non-linear activation function, such as tanh, and $\tilde{\h}_s\in \mathbb{R}^H$ is an exponentially smoothed version of the hidden state $\hat{\h}_s$. When the output is continuous, the output from the final hidden state is given by:
\be
\hat{\y}_{t+m}= \W_y\hat{\h}_t + \bias_y,
\ee
with the starting condition in each sequence,
$\hat{\h}_{t-p+1}=\act(\W_h\x_{t-p+1})$. 

\section{Times Series Modeling}\label{sect:tsm_intro}
This section bridges the time series modeling literature \citep{boxjen76,UBMA_276336909,doi:10.1111/jtsa.12464} and the machine learning literature.
We shall assume here for ease of exposition that the time series data is univariate ($d=1$), $\mathcal{D}:=\{y_t\}_{t=1}^N$, and thus the predictor is endogenous. 

Since autoregressive ($AR(p)$) models are well known in time series modeling, we find it instructive to show that plain RNNs are non-linear AR(p) models with drift $\mu$ and coefficients $\{\phi_i\}_{i=1}^p$.  For ease of exposition, consider the simplest case of a RNN with one hidden unit, $H=1$. Without loss of generality, we set $U_h=W_h=\phi\in\mathbb{R}$, $W_y=1$, $b_h=0$ and $b_y=\mu\in\mathbb{R}$. Then we can show by backward substitution that a plain-RNN, $F_{W, b}(\bm{\underbar{x}}_t)$, with sequence length $p$, is a non-linear auto-regressive, $NAR(p)$, model of order $p$ as follows:
\begin{eqnarray*}
\hat{h}_{t-p+1} &=& \act(\phi y_{t-p+1}),\\
\hat{h}_{t-p+2} &=& \act(\phi\hat{h}_{t-p+1} + \phi y_{t-p+2}),\\
\dots &=&\dots,\\
\hat{h}_{t} &=& \act(\phi \hat{h}_{t-1} +\phi y_{t}),\\
\hat{y}_{t+m} &=& \hat{h}_{t} + \mu,
\end{eqnarray*}
to yield the non-linear autoregressive model
\be
\hat{y}_{t+m}=\mu + \act(\phi(1 + \act(\phi(L+ \act(\phi(L^2 + \dots +  \act(\phi L^{p-1})\dots)[y_t].
\ee
When the activation is the identity function $\act:=I_d$, then we recover the AR(p) model
\be
\hat{y}_{t+m}= \mu + \sum_{i=0}^{p-1}\phi_{i+1} L^i[y_{t}],
\ee
where $\phi_i=\phi^{i}$ and thus these autoregressive coefficients geometrically decay with increasing lag when $|\phi| < 1$.

\paragraph{$\alpha$-RNNs} The $\alpha$-RNN(p) is almost identical to a plain RNN, but with an additional scalar smoothing parameter, $\alpha$, which provides the recurrent network with ``long-memory'', i.e. autoregressive memory beyond the sequence length. For avoidance of doubt, we are not suggesting that the $\alpha$-RNN has an additional cellular memory, as in LSTMs. To see this, let us consider a one-step ahead univariate $\alpha$-RNN(p) in which the smoothing parameter is fixed. For each time step $s=t-p+2,\dots,t$:
\begin{eqnarray*}
(\textrm{output})&~& \hat{y}_{t+1}=W_y \hat{h}_t + b_y,\\
(\textrm{hidden state update})&~& \hat{h}_s=\act(U_h\tilde{h}_{s-1} + W_hy_s + b_h),\\
(\textrm{smoothing})&~& \tilde{h}_{s} = \alpha \hat{h}_{s} + (1-\alpha)\tilde{h}_{s-1}.
\end{eqnarray*}
This model augments the plain-RNN by replacing $\hat{h}_{s-1}$ in the hidden layer with an exponentially smoothed hidden state $\tilde{h}_{s-1}$. The effect of the smoothing is to provide infinite memory when $\alpha\neq 1$. For the special case when $\alpha=1$, we recover the plain RNN with short memory of length $p<<N$.

We can easily see this informally by simplifying the parameterization and considering the unactivated case. Setting $b_y=b_h=0$,  $U_h=W_h=\phi\in\mathbb{R}$ and $W_y=1$:
\begin{eqnarray}
\hat{y}_{t+1}&=&\hat{h}_t,\\
&=&\phi(\tilde{h}_{t-1} + y_t),\\
&=&\phi(\alpha \hat{h}_{t-1} + (1-\alpha)\tilde{h}_{t-2}  + y_t),
\end{eqnarray}
with the starting condition in each sequence, $\hat{h}_{t-p+1}=\phi y_{t-p+1}$. 
With out loss of generality, consider $p=2$ lags in the model so that $\hat{h}_{t-1}=\phi y_{t-1}$. Then
\be
\hat{h}_t=\phi(\alpha \phi y_{t-1} + (1-\alpha)\tilde{h}_{t-2}  + y_t)
\ee
and the model can be written in the simpler form
\be\label{eq:alpha-mem}
\hat{y}_{t+1}= \phi_1y_t + \phi_2 y_{t-1}  + \phi (1-\alpha)\tilde{h}_{t-2}, 
\ee
with auto-regressive weights $\phi_1:=\phi$ and $\phi_2:=\alpha\phi^2$. We now see that there is a third term on the RHS of Eq. \ref{eq:alpha-mem} which vanishes when $\alpha=1$ but otherwise provides infinite memory to the model since $\tilde{h}_{t-2}$ depends on $y_1$, the first observation in the entire time series, not just the first observation in the sequence. To see this, we unroll the recursion relation in the exponential smoother:
\be
\tilde{h}_{t+1}= \alpha\sum_{s=0}^{t-1}(1-\alpha)^s \hat{h}_{t-s} + (1-\alpha)^ty_{1},
\ee
where we used the property that $\tilde{h}_1=y_1$. It is often convenient to characterize exponential smoothing by the \textbf{half-life}--- the number of lags needed for the coefficient $(1-\alpha)^s$ to equal a half, which is $s=-1/log_2(1-\alpha)$. To gain further insight we use partial auto-correlations to characterize the memory of the model.

\subsection{Partial Autocorrelation}
We consider autoregressive time series models, with additive white noise of the form 
$$y_{t}=\hat{y}_{t} + \epsilon_t, ~ \epsilon_t\sim N(0,\sigma_n^2),$$
which carry a signature which allows its order, $p$, to be determined from ``covariance stationary'' time series data (see supplementary material for a definition of covariance stationarity). This signature encodes the memory in the model and is given by ``partial autocorrelations''. Informally each partial autocorrelation measures the correlation of a random variable, $y_t$, with its $h^{th}$ lag, $y_{t-h}$, while controlling for intermediate lags. The partial autocorrelation must be non-zero to be able to predict $y_t$ from $y_{t-h}$. The sign of the partial autocorrelation is also useful for \textbf{interpretability} and describes the directional relationship between the random variables. The formal definition of the partial autocorrelation is now given.

\begin{defn}[Partial Autocorrelation]
A partial autocorrelation at lag $h\geq 2$ is a conditional autocorrelation between a variable, $y_t$, and its $h^{th}$ lag, $y_{t-h}$ under the assumption that the values of the intermediate lags, $y_{t-1},\dots, y_{t-h+1}$ are controlled:

$$\tilde{\tau}_h:=\tilde\tau_{t,t-h}\\
:=\frac{\tilde{\gamma}_{h}}{\sqrt{\tilde\gamma_{t,h}}\sqrt{\tilde\gamma_{t-h,h}}},$$

where 
\begin{eqnarray*}
\tilde\gamma_h&:=&\tilde\gamma_{t,t-h}\\
&:=&\mathbb{E}[y_t-P(y_t \given y_{t-1},\dots, y_{t-h+1}), \\
&& y_{t-h} -P(y_{t-h} \given y_{t-1},\dots, y_{t-h+1})]
\end{eqnarray*}
is the lag-$h$ partial autocovariance, $P(W \given Z)$ is an orthogonal projection of W onto the set $Z$ and 
\be
\tilde\gamma_{t,h}:= \mathbb{E}[(y_t-P(y_t \given y_{t-1},\dots, y_{t-h+1}))^2]. 
\ee
The partial autocorrelation function (PACF) $\tilde{\tau}_h:\mathbb{N} \rightarrow [-1,1]$ is a map $h:\mapsto \tilde{\tau}_h$. The plot of $\tilde{\tau}_h$ against $h$ is referred to as the \bf{partial correlogram}.
\end{defn}

The PACF of the RNN(p) can be used to determine the lower bound on the sequence length in an $\alpha$-RNN(p). To see this, we first show that the partial autocorrelation of the $\alpha$-RNN(p) is time independent and has a sharp cut-off after $p$ lags if $\alpha=1$. 
\\
\begin{thm}[Partial autocorrelation of an $\alpha$-RNN(p)] \label{thm:PA}
The partial autocorrelation of the $\alpha$-RNN(p) is time independent and exhibits a cut-off after $p$ lags: $\tilde{\tau}_s=0, s>p$ if $\alpha=1$. If $\alpha\in(0,1)$ the $\alpha$-RNN(p) has non-zero partial autocorrelations at lags beyond the sequence length.
\end{thm}
\vspace{0.5cm}
\begin{remark}
The partial autocorrelation can be used to identify the order of the RNN model from the estimated PACF and hence determines the sequence length in the $\alpha$-RNN which is guaranteed to have at least the same order for $\alpha\in(0,1]$.
\end{remark}

\begin{figure}[h!]
\centering
\includegraphics[width=0.5\columnwidth]{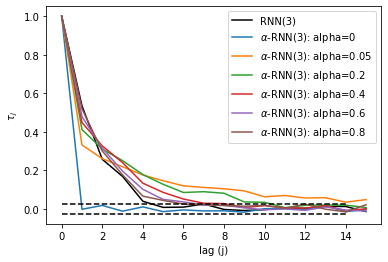}
\caption{\textit{The fitted partial correlogram of univariate data generated by various additive noise $\alpha$-RNN(3) models.}}
\label{fig:pacf}
\end{figure}

See Appendix B for the proof. For $\alpha \in (0,1)$, the $\alpha$-RNN has non-zero partial autocorrelations $\tilde{\tau}_s\neq 0, s>p$. It is easy to see this from the additional term containing $\alpha$ in Equation 
\ref{eq:alpha-mem}. Further insight can be gained from Figure \ref{fig:pacf} which shows the fitted partial correlogram from data generated by an $\alpha$-RNN(3) with additive white noise. We observe that the memory is always longer than the sequence length of 3 when $\alpha\in(0,1)$. As $\alpha$ approaches zero, the model has increasing memory, although the model has no memory in the limit $\alpha=0$.  The theorem and the properties of $\alpha$-RNN(p) suggest determining the sequence length $p$ from the fitted PACF of each covariate. Moreover, the prediction horizon should not exceed that suggested by the maximum order of statistically significant partial autocorrelations.

\subsection{Stability}
Times series modeling also places emphasis on model "stability" --- this is the model attribute that past random disturbances decay in the model and the effect of lagged data becomes less relevant to the model output with increasing lag. We present the following theorem which shows that the stability of the $\alpha$-RNN model is determined solely by hidden state's activation function, with the property that $|\act(\cdot)|< 1$.

\begin{thm}[Stability of RNN(p) models] \label{thm:stability}
Suppose that there exists an invertible nonlinear function of the lag operator $\Phi(L)$ of the form:
\begin{eqnarray*}
y_t&=&\Phi^{-1}(L)[\epsilon_t]\\
&=&(1-\act(\phi\act(\phi\act(\phi\act(\dots)) + \dots \\
&+&  \phi L^2) + \phi L)^{-1}[\epsilon_t],
\end{eqnarray*}
where, without loss of generality, we have again conveniently set $W_x=U_h=\phi$, $W_y=1$ and $b_h=b_y=1$.
Then the RNN is stable if and only if  $|\act(x)|< 1$ for all finite $x$.
\end{thm}
See Appendix B for the proof. In particular, the theorem justifies the common choice of $\act(\cdot):=tanh(\cdot)$ and while sigmoid is also another viable choice, it is too restrictive as it prohibits negative partial auto-correlations. We shall see in Section \ref{sect:results} that negative partial auto-correlations arise in time series data.

\subsection{Vanishing gradient}
It is well known that plain-RNNs exhibit a vanishing gradient problem \citep{pascanu2012difficulty, bayer}. Following \cite{pascanu2012difficulty}, we extend some of the analysis of BPTT to $\alpha$-RNNs. The $\alpha$-RNN(p) BPTT can be written as:
\begin{eqnarray}\label{eq:loss_gradient}
\frac{\partial\mathcal{L}}{\partial W}&=&\sum_{t=1}^N\frac{\partial\mathcal{L}}{\partial W}\\
&=& \sum_{t=1}^N\frac{\partial\mathcal{L}}{\partial \tilde{h}_t}\sum_{k=t-p}^{t-1}\prod_{i=k+1}^{t-1} \frac{\partial \tilde{h}_i}{\partial \tilde{h}_{i-1}} \frac{\partial \tilde{h}_k}{\partial W}
\end{eqnarray}
for some generic loss function $\mathcal{L}_t$, where 
\begin{eqnarray*}
\frac{\partial \tilde{h}_i}{\partial \tilde{h}_{i-1}}&=&(1-\alpha) + \alpha\frac{\partial \hat{h}_i}{\partial \tilde{h}_{i-1}}\\
&=& (1-\alpha) + \alpha\sigma'(W_hx_i + U_h\tilde{h}_{i-1} +b_h)U_h.
\end{eqnarray*}
Substituting the expression for $\frac{\partial \tilde{h}_i}{\partial \tilde{h}_{i-1}}$ into Equation \ref{eq:loss_gradient} gives an expression proportional to 
\begin{equation*}
\prod_{i=k+1}^{t-1}(1-\alpha) + \alpha\sigma'(I_i)U_h, ~I_i:=W_hx_i + U_h\tilde{h}_{i-1} +b_h.
\end{equation*}
When $\alpha=1$ this expression is $\prod_{i=k+1}^{t-1}\sigma'(I_i)U_h$. When $\sigma(\cdot):=tanh(\cdot)$, this product goes to zero with increasing function compositions due to the gradient of tanh vanishing with large $|x|$ and the product of small functions. However, when $\alpha \neq 1$, the additional term provides an additional contribution to the gradient which is non trivial for small $\alpha$ and independent of the input.



Thus the $\alpha$-RNN will not only alleviate the vanishing gradient problem, has guaranteed stability if we choose a tanh activation function, but exhibits non-zero partial auto-correlations up to at least lag $p$ for $\alpha\in (0,1]$. 

The $\alpha$-RNN model can be trained by treating $\alpha$ as an additional parameter to be fitted with stochastic gradient descent. The choice to pre-determine that $\alpha$ is independent of time is obviously limited to stationary time series. While this is restrictive, it suggests that a simple statistical test of the time series can pre-determine the efficacy of the approach. Moreover, if the data is covariance stationary, then the $\alpha$-RNN will preserve the stationarity of the partial auto-correlation structure, eliminating the need for more complex architectures such as GRUs and LSTMs (which are often motivated purely on the basis of the vanishing gradient problem). Such a procedure shall be demonstrated in Section \ref{sect:results}. We can extend the model to non-stationary time series, which exhibit dynamic partial autocorrelation, by using a dynamic version of exponential smoothing.

\section{Dynamic $\alpha_t$-RNNs} \label{sect:alpha_t}
The extension of RNNs to dynamical time series models, suitable for non-stationary time series data, relies on dynamic exponential smoothing is a time dependent, convex, combination of the smoothed output, $\tilde{h}_t$, and the hidden state $\hat{h}_{t}$:
\be
\tilde{h}_{t+1} = \alpha_{t}\hat{h}_{t} + (1-\alpha_t)\tilde{h}_{t},
\ee
where $\alpha_t\in [0,1]$ denotes the dynamic smoothing factor which can be equivalently written in the one-step-ahead forecast of the form
\be 
\tilde{h}_{t+1} = \tilde{h}_{t} + \alpha_t ( \hat{h}_t - \tilde{h}_{t}).
\ee 
Hence the smoothing can be viewed as a form of dynamic forecast error correction. When $\alpha_t=0$, the forecast error is ignored and the smoothing merely repeats the current hidden state $\tilde{h}_t$, to the effect of the model losing its memory. When $\alpha_t=1$, the forecast error overwrites the current hidden state $\tilde{h}_t$. The smoothing can also be viewed a weighted sum of the lagged observations, with lower or equal weights, $\alpha_{t-s}\prod^s_{r=1}(1-\alpha_{t-r+1})$ at the $s\geq 1$ lagged hidden state, $\hat{h}_{t-s}$:
\begin{equation*}
\tilde{h}_{t+1}= \alpha_t \hat{h}_t +\sum_{s=1}^{t-1}\alpha_{t-s}\prod^{s}_{r=1}(1-\alpha_{t-r+1})\hat{h}_{t-s}+ g(\mathbf{\alpha}),
\end{equation*}
where $g(\mathbf{\alpha}):=\prod^{t-1}_{r=0}(1-\alpha_{t-r})\tilde{y}_{1}$. Note that for any $\alpha_{t-r+1}=1$, the smoothed hidden state $\tilde{h}_{t+1}$ will have no dependency on all lagged hidden states $\{\hat{h}_{t-s}\}_{s \geq r}$. The model simply forgets the hidden states at or beyond the $r^{th}$ lag.

\subsection{Neural network exponential smoothing}
While the class of $\alpha_t$-RNN models under consideration is free to define how $\alpha$ is updated (including changing the frequency of the update) based on the hidden state and input, a convenient choice is use a recurrent layer. Returning again to the more general setup with a hidden state vector $\hat{\h}_t\in \mathbb{R}^H$, let us model the smoothing parameter $\hat{\bm{\alpha}}_t \in [0,1]^H$ to give a filtered time series
\be\label{eq:smoothing}
\tilde{\h}_t = \hat{\bm{\alpha}}_t\circ \hat{\h}_t + (1-\hat{\bm{\alpha}}_t)\circ\tilde{\h}_{t-1},
\ee
where $\circ$ denotes the Hadamard product between vectors. This smoothing is a vectorized form of the above classical setting, only here we note that when $(\hat{\bm{\alpha}}_t)_i=1$, the $i^{th}$ component of the hidden variable is unmodified and the past filtered hidden variable is forgotten. On the other hand, when the $(\hat{\bm{\alpha}}_t)_i=0$, the $i^{th}$ component of the hidden variable is obsolete, instead setting the current filtered hidden variable to its past value.  The smoothing in Equation \ref{eq:smoothing} can be viewed then as updating long-term memory, maintaining a smoothed hidden state variable as the memory through a convex combination of the current hidden variable and the previous smoothed hidden variable. 

The hidden variable is given by the semi-affine transformation:
\be \label{eq:hidden_state}
\hat{\h}_t =\act(\U_{h}\tilde{\h}_{t-1} + \W_{h}\x_t+\bias_{h})
\ee
which in turns depends on the previous smoothed hidden variable. Substituting Equation \ref{eq:hidden_state} into Equation \ref{eq:smoothing} gives a function of $\tilde{\h}_{t-1}$ and $\x_t$:
\begin{eqnarray}\label{eq:smoothing_hidden _state}
\tilde{\h}_t &=& g(\tilde{\h}_{t-1}, \x_t; \bm{\alpha})\\
&=& \hat{\bm{\alpha}}_t \circ \act(\U_{h}\tilde{\h}_{t-1} + \W_{h}x_t+\bias_{h}) + (1-\hat{\bm{\alpha}}_t)\circ\tilde{\h}_{t-1}.
\end{eqnarray}
We see that when $\hat{\bm{\alpha}}_t=\bm{0}$, the smoothed hidden variable $\tilde{\h}_t$ is not updated by the input $\x_t$. Conversely, when $\hat{\bm{\alpha}}_t=\bm{1}$, we observe that the hidden variable locally  behaves like a non-linear autoregressive series. Thus the smoothing parameter can be viewed as the sensitivity of the smoothed hidden state to the input $\x_t$. 

The challenge becomes how to determine dynamically how much error correction is needed. As in GRUs and LSTMs (see Appendix D for further details), we can address this problem by learning $\hat{\bm{\alpha}}=F_{(W_{\alpha}, U_{\alpha}, b_{\alpha})}(\bm{\underbar{x}}_t)$ from the input variables with the recurrent layer parameterized by weights and biases $(\W_{\alpha}, \U_{\alpha}, \bias_{\alpha})$ so that
\be
\hat{\bm{\alpha}}_t=\sigma_s(\U_{\alpha}\tilde{\h}_{t-1} + \W_\alpha\x_t + \bias_\alpha),
\ee
 where $\sigma_s(x):=\frac{1}{1+e^{-x}}$. Again, the one-step ahead forecast of the smoothed hidden state, $\tilde{\h}_t$, is the filtered output of another plain RNN with weights and biases $(\W_h,\U_h, \bias_h)$. 

\begin{figure}[h!]
\centering
\begin{tabular}{cc}
\includegraphics[width=0.52\columnwidth]{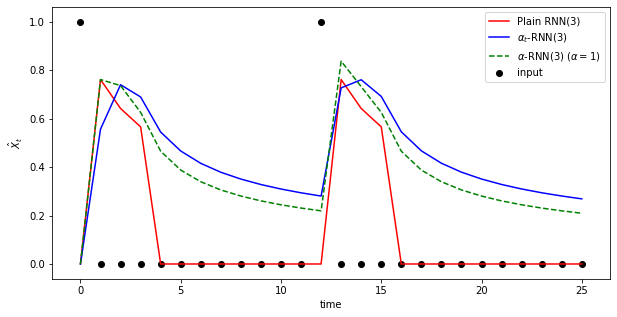} &
\includegraphics[width=0.4\columnwidth]{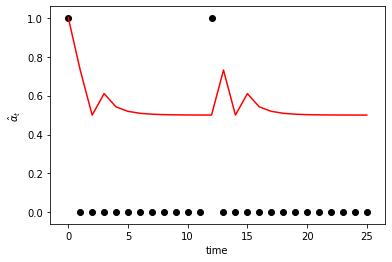}\\
(a) & (b) 
\end{tabular}
\caption{\textit{An illustrative example of the response of an $\alpha$-RNN in comparison with a plain-RNN. The plain-RNN is an Elman network without smoothing and corresponds to the neural network of recurrent cells shown in Figure 1a. (a) Comparison of model responses in the presence of shocks to the inputs. (b) Response of $\hat{\alpha}_t$ to shocks in the input. The RNN model is chosen for illustrative purposes to be a RNN(3) model, i.e. with a sequence length of 3.}}
\label{fig:toy_gru}
\end{figure}

\paragraph{Comparison with $\alpha$-RNNs} Figure \ref{fig:toy_gru}(a) shows the response of a univariate $\alpha$-RNN model when the input consists of two unit impulses and zeros otherwise. For simplicity, the sequence length is assumed to be $3$ (i.e. the RNN has a memory of 3 lags), the biases are set to zero, all the weights are set to one and $\act(x):=tanh(x)$. Note that the weights have not been fitted here, we are merely observing the effect of smoothing on the hidden state for the simplest choice of parameter values. The RNN loses memory of the unit impulse after three lags, whereas the RNNs with smooth hidden states maintain memory of the first unit impulse even when the second unit impulse arrives.  The figure also shows an $\alpha_t$-RNN model, which although appears insignificant in this example, allows the model to fit non-stationary time series data. This is because the time dependent $\alpha_t$ results in dynamic partial autocorrelations. The response of $\hat{\alpha}_t$ to shocks can be seen in Figure \ref{fig:toy_gru}(b). Note that the value of $\alpha$ is initially set to one since no long-term memory is needed. Hence the response to the second shock at time $t=12$ is more insightful.


\section{Uncertainty Quantification}\label{sect:bayesian}

Uncertainty refers to the statistically unmeasurable situation of Knightian uncertainty, where the event space is known but the probabilities are not \citep{sowers2015}.  Oftentimes, a forecast may be shrouded in uncertainty arising from noisy data or model uncertainty, either through incorrect modeling assumptions or parameter error. It is desirable to characterize this uncertainty in the forecast. In conventional Bayesian modeling,  uncertainty is used to learn from small amounts of low dimensional data under parametric assumptions on the prior. The choice of the prior is typically the point of contention and chosen for solution tractability rather than modeling fidelity. Recently, deterministic deep learners have been shown to scale well to large, high dimensional, datasets. However, the probability vector obtained from a classification network is often erroneously interpreted as model confidence  \citep{gal2016uncertainty}.

Following \cite{polson2017}, we estimate the aleatoric uncertainty by treating the neural network parameters (weights and biases) as random variables. Priors are placed over parameters and an approximation of the posterior is estimated by "variational inference", an approximate technique which allows multi-modal likelihood functions to be extremized with a standard stochastic gradient descent algorithm. See Appendix E for details of this approach. An alternative to variational algorithms was recently proposed by \cite{gal2016uncertainty} and builds on an efficient dropout regularization technique in a frequentist framework, but it not explored here as the uncertainty estimates are highly sensitive to the dropout parameter.

We use a scale mixture of two zero mean Gaussian densities as the prior. Our prior resembles a spike-and-slab prior where all the prior parameters are shared among all the weights. Once the network has been fitted, the weights and biases are bootstrap sampled from the variational posterior and the model is used to generate a corresponding prediction distribution at each time $t$ in the sequence. The sample mean and standard deviation are used to give the boostrapped confidence intervals.

\section{Numerical Experiments} \label{sect:results}
This section describes numerical experiments using time series data to evaluate the various RNN models. Unless stated otherwise, all models are implemented in v1.15.0 of TensorFlow \cite{abadi2016tensorflow}. Times series cross-validation is performed using separate training, validation and test sets. To preserve the time structure of the data and avoid look ahead bias, each set represents a contiguous sampling period with the test set containing the most recent
observations.  

To prepare the training, validation and testing sets for m-step ahead prediction, we shall either use "direct" forecasting or "rolling" forecasts. The former sets the target variables (responses) to the $t+m$ observation, $y_{t+m}$, and use the lags from $t-p+1, \dots t$ for each input sequence. This is repeated by incrementing $t$ until the end of each test set. The latter rolls one period forecasts out to the $t+m$ observation, generating and subsequently updating the input with intermediate forecasts. The latter is typically used in ARIMA methods and is more appropriate when the prediction horizon is longer than the sequence length in the model (i.e. $p$ in the case of the ARIMA model). The former is faster and better suited to capturing longer-term trends and seasonal effects. In our experiments, each element in the input sequence is either a scalar or vector and the target variables are scalar.


We use the \verb|SimpleRNN| Keras method with the default settings to implement a fully connected RNN. Tanh activation functions are used for the hidden layer and, unless otherwise stated, the number of units found by time series cross-validation with five folds $H\in\{5,10,20\}$ and $L_1$ regularization, with $\lambda_1\in\{0, 10^{-3},10^{-2}\}$. The Glorot and
Bengio uniform method \citep{Glorot2010} is used to initialize the non-recurrent weight matrices and, for stability, an orthogonal matrix is used to initialize the recurrence weights to ensure that the absolute value of the eigenvalues are initially bounded by unity \citep{henaff2017recurrent}.  Keras's \verb|GRU| method is implemented using version 1406.1078v, which applies the reset gate to the hidden state before matrix multiplication. Similarly, the \verb|LSTM| method in Keras is used.  Tanh activation functions are used for the recurrence layer and sigmoid activation functions are used for all other gates. We implement an exponential smoothing (ES) layer in Keras and create a \verb|ESRNN| model by combining a simple RNN with the ES layer. Note that \cite{SMYL202075} consider more complex variants of the simple RNN, stacking multiple recurrent layers and using dilation and dual attention to capture multi-scale effects and a technique suitable for high dimensional datasets respectively. 

However, such variants could equally be combined with any of the recurrent networks considered and thus, for the purpose of benchmarking and understanding the effect of the exponential smoothing layer, we do not consider these additional architectural complexities here. Moreover, it is much less clear how the theory developed in this paper is applicable in the presence of such complexities. 

Unless otherwise stated, we shall use the following parameters for experiments. The loss function is chosen to be the mean squared error with an additional LASSO penalty term, an $L_1$ norm of the weights scaled by a regularization parameter $\lambda_1$. 

Each architecture is trained for up to 2000 epochs with an Adam optimization algorithm with default parameter values and using a mini-batch size of 1000 drawn from the training set. Early stopping is used with a "patience" of 50 to 100 and minimum absolute difference between $10^{-8}$ and $10^{-6}$. Patience defines the number of iterations that must consecutively satisfy the threshold constraint on the absolute difference between current and previous losses. So, for example, if the patience is set to 50 and the minimum loss difference is $10^{-8}$, then fifty consecutive loss evaluations on mini-batch updates must each be within $10^{-8}$ of the previous update before the training terminates.  In practice, the actual number of epochs required varies between each training due to the randomization of the weights and biases, and across different architectures and is typically between 200 and 1500. The 2000 epoch limit is chosen as it provides an upper limit which avoids excessive training times. No randomization is used in the mini-batching sampling in order to preserve the ordering of the data.

To evaluate the forecasting accuracy, we set the forecast horizon to up to ten steps ahead instead of the usual step ahead forecasts often presented in the machine learning literature --- longer forecasting horizons are often more relevant due to operational constraints in industry applications and are more challenging when the data is non-stationary since the network's fixed partial auto-correlation will not adequately capture the observed changing autoregressive structure. The numerical results are separated in two areas of interest: (i) properties of recurrent architectures on archetypal data properties such as seasonality and trends; and (ii) evaluation on real data to demonstrate their utility in industrial forecasting.

\subsection{LLM Seasonality DGP} \label{sect:results_llm}
To characterize the ability of RNNs to directly capture seasonality and trends from noisy data, without the need to separately detrend and deseasonalize the data, we generate hourly data from an additive local level model with daily seasonality and iid noise \cite{harvey1990forecasting}:
\begin{eqnarray*}
\textrm{observed series}: y_t &=& \mu_t + \gamma_t + \epsilon_t,~\epsilon_t\sim \mathcal{N}(0,\sigma_u^2),\\
\textrm{latent level}: \mu_t &=& \mu_{t-1} + \chi_t,~ \chi_t\sim \mathcal{N}(0,\sigma_\chi^2),\\ 
\textrm{latent seasonal}: \gamma_t &=& \sum_{j=1}^{s-1} -\gamma_{t-j} + \omega_t,~\omega_t\sim \mathcal{N}(0,\sigma_\omega^2),
\end{eqnarray*}
for $t\in\{s,\dots, N\}$. Choosing $s=24$, we simulate $N=10,000$ observations under noise variances $\sigma_u^2=300, \sigma_\chi^2 = 1, \sigma_\omega^2 = 1$. The first $8,000$ observations are used for training and the remaining are used for testing. 

The data is non-stationary --- we accept the Null hypothesis of the augmented Dickey-Fuller test which states that the AR model contains a unit root \citep{doi:10.1080/07350015.1995.10524601}. The test statistic is  $-2.51$ and the p-value is $0.114$ (the critical values are 1\%: -3.431, 5\%: -2.862, and 10\%: -2.567).
We choose a sequence length of $p=30$ which is greater than the periodicity of the data. As a baseline for the recurrent networks, we fit an ARIMA model to a subset of the most recent 1000 observations in the training set. Because the data is non-stationary, we first use a seasonal trend decomposition and select a model for forecasting the stationary residual series. Since the focus of this paper is assessing the performance of $\alpha$-RNNs over multi-period horizons, we configure the ARIMA model as a rolling five-step ahead forecast. This requires prediction and subsequent use of intermediate lags $\hat{y}_{t+1}, \dots, \hat{y}_{t+4}$. We find the optimal coefficient $p=q=2$ based on the in-sample AIC. 

RNNs permit a more flexible forecasting configuration - the five-step ahead forecast is directly given by the model without the need for predicting the intermediate forecasts $\hat{y}_{t+1}, \dots, \hat{y}_{t+4}$. Moreover, for some architectures, we observe that there is merit in directly forecasting without the need for seasonal and trend decomposition. This aspect is of interest more generally when seasonal and trend decomposition is inadequate, e.g. if the trend is correlated with the seasonal component or the periodicity of the seasonal component is not constant. There is also a fashion towards end-to-end forecasting without the need for decomposition based preprocessing techniques.

\begin{figure}[H]
\begin{subfigure}[t]{0.3\columnwidth}
\centering
\includegraphics[width=\textwidth]{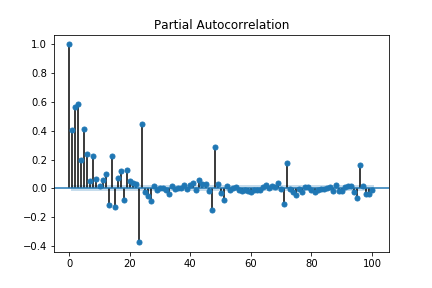} 
\caption{Observed}
\end{subfigure}
\begin{subfigure}[t]{0.3\columnwidth}
\centering
\includegraphics[width=\textwidth]{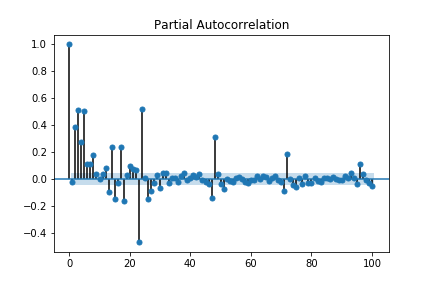} 
\caption{ARIMA+decomp)}
\end{subfigure}
\begin{subfigure}[t]{0.3\columnwidth}
\centering
\includegraphics[width=\textwidth]{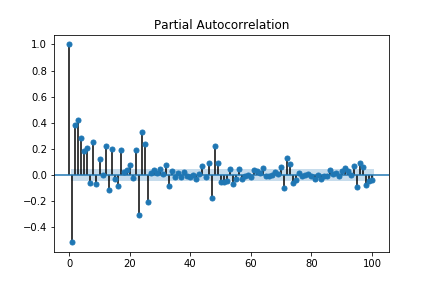}
\caption{Simple RNN}
\end{subfigure}
\newline
\begin{subfigure}[t]{0.3\columnwidth}
\centering
\includegraphics[width=\textwidth]{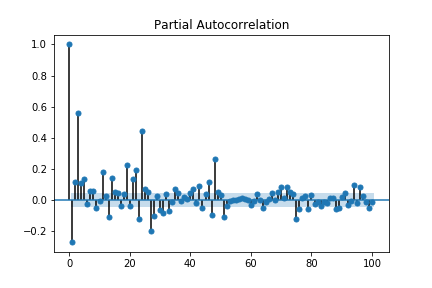}
\caption{ES-RNN}
\end{subfigure}
\begin{subfigure}[t]{0.3\columnwidth}
\centering
\includegraphics[width=\textwidth]{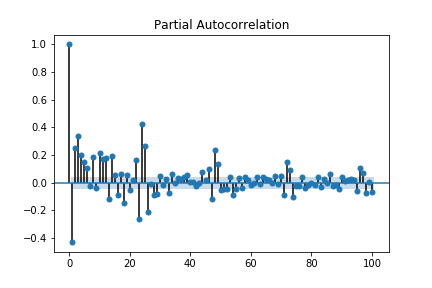} 
\caption{$\alpha$-RNN}
\end{subfigure}
\begin{subfigure}[t]{0.3\columnwidth}
\centering
\includegraphics[width=\textwidth]{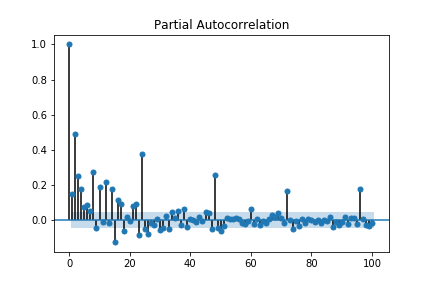} 
\caption{$\alpha_t$-RNN}
\end{subfigure}
\newline
\begin{subfigure}[t]{0.3\columnwidth}
\centering
\includegraphics[width=\textwidth]{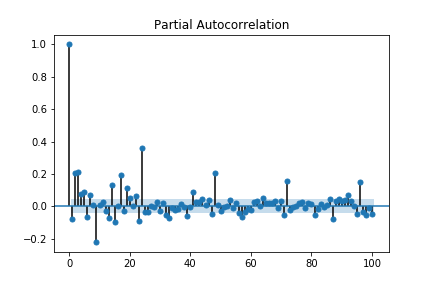} 
\caption{GRU}
\end{subfigure}
\begin{subfigure}[t]{0.3\columnwidth}
\centering
\includegraphics[width=\textwidth]{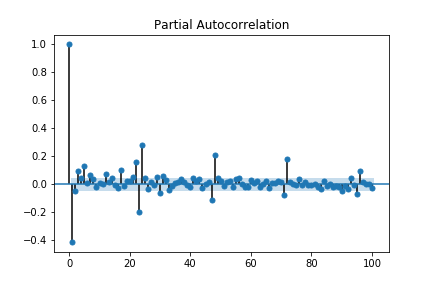} 
\caption{LSTM}
\end{subfigure}

\caption{\textit{The PACFs are compared between a 5-step seasonality decomposed ARIMA model and various recurrent architectures.}}
\label{fig:sea2}
\end{figure}

In all subsequent figures and tables, forecasting results are reported against the original time series. Figure \ref{fig:sea2} compares the PACFs of various models on the test set. Figure \ref{fig:sea2}(a) shows the PACF on the generated test data  --- the positive lags at 24, 48, 72 and 96, due to the seasonality, are clearly shown. Figure \ref{fig:sea2}(b) shows the PACF of the five-step ahead forecast from the ARIMA(2,0,2) model, where seasonal and trend decomposition is used. Since the data is generated by a process which is well suited to seasonal and trend decomposition, we observe as expected that the seasonally decomposed ARIMA provides an adequate baseline for model performance and hence its PACF is close to the PACF of the observed test set.  Figures \ref{fig:sea2}(c)-(h) compare the PACF of the various recurrent network forecasts, \emph{without} the use of seasonal decomposition. Since the data is non-stationary, we would expect limitations with using architecture with fixed partial auto-correlations. Indeed, a spurious large and negative lag 1 partial autocorrelation is observed in the PACF of the Simple RNN, the ES-RNN, and the $\alpha$-RNN. Spurious low order lagged partial autocorrelations are especially problematic for prediction if they are the wrong sign. In the case of the former two, we further see evidence of inability to resolve the periodic structure at larger lags. In contrast, the $\alpha_t$-RNN, the GRU, and the LSTM not only resolve the periodic structure at larger lags, but more importantly, the former two do not exhibit the large spurious lag-1 negative partial autocorrelation.

Figure \ref{fig:sea2} illustrates the ability of $\alpha_t$-RNNs and GRUs to intrinsically capture seasonality and trends without introducing spurious low-order lags. However, the comparison with ARIMA must factor for a number of differences in the design of the experiment. In the next experiment, we put the ARIMA and recurrent architectures on an equal footing and configure the experiments identically. Seasonal and trend decomposition is applied first, and five-step ahead rolling forecasts are then performed by all methods (requiring generation of intermediate time steps). In this way, ARIMA can be fairly compared with the various RNNs, and since the resulting residual time series is stationary, we expect to see the best performance from methods designed for stationary times series, i.e. RNNs, ES-RNNs, and $\alpha$-RNNs.

Figure \ref{fig:dcp}(a)-(g) compares the model forecast over a horizon of 1000 observations. The errors are comparable in size and under 60, except for the GRU which has errors above 60 but under 100. We note that all methods typically forecast more moderate values than the observed series. Figure \ref{fig:dcp}(h) compares the forecasts against the observed test set (black line) over a shorter horizon of 100. The RMSEs are provided in the parentheses within the legend and the color scheme for the forecast in \ref{fig:dcp}(a)-(g) is consistent with Figure \ref{fig:dcp}(h). While the observations are noisy and hence any gains across different architectures are more difficult to visualize, we observe that $\alpha$-RNN (green lines) tends to more closely match the observed and the GRU (yellow) exhibits more spurious over and undershoots.

\begin{figure}[H]
\begin{subfigure}[t]{0.31\columnwidth}
\centering
\includegraphics[width=\textwidth, height= 0.2\textheight]{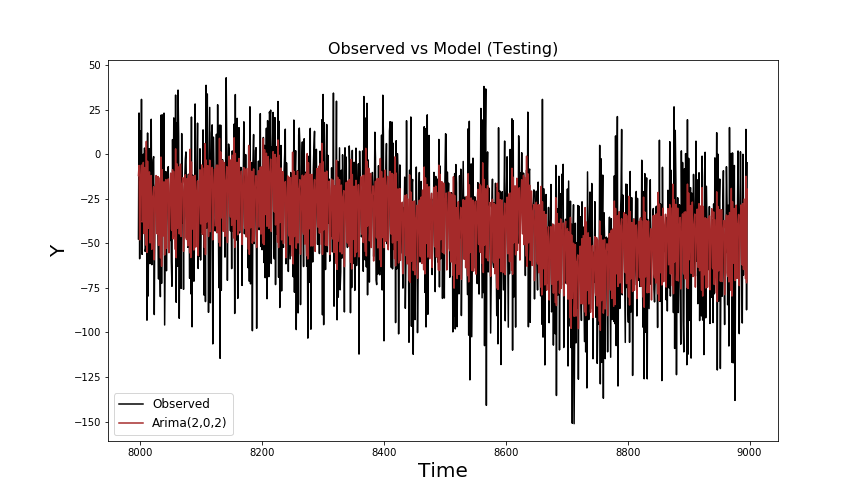} 
\caption{ARIMA(2,0,0).}
\end{subfigure}
\begin{subfigure}[t]{0.31\columnwidth}
\centering
\includegraphics[width=\textwidth, height=0.2\textheight]{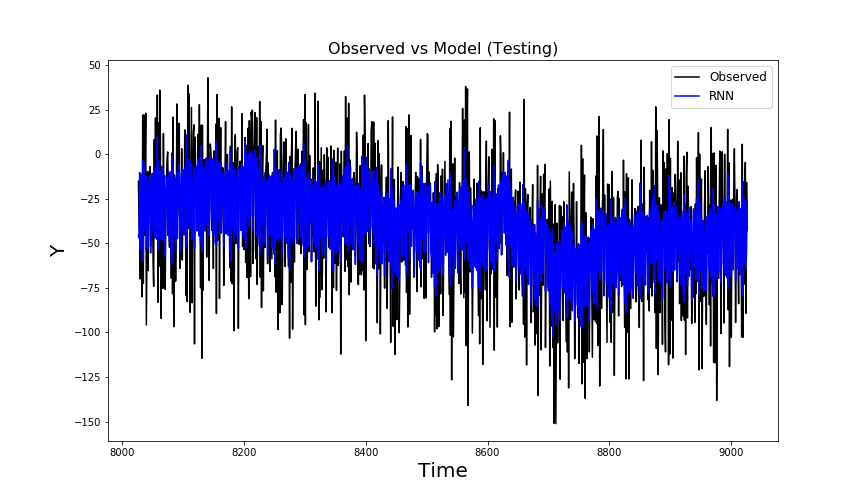} 
\caption{RNN.}
\end{subfigure}
\begin{subfigure}[t]{0.31\columnwidth}
\centering
\includegraphics[width=\textwidth, height= 0.2\textheight]{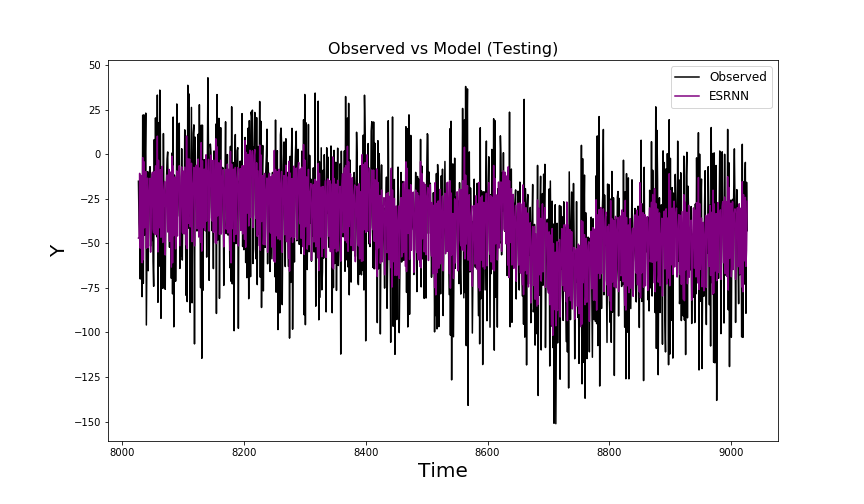} 
\caption{ES-RNN.}
\end{subfigure}
\newline
\begin{subfigure}[t]{0.31\columnwidth}
\centering
\includegraphics[width=\textwidth, height= 0.2\textheight]{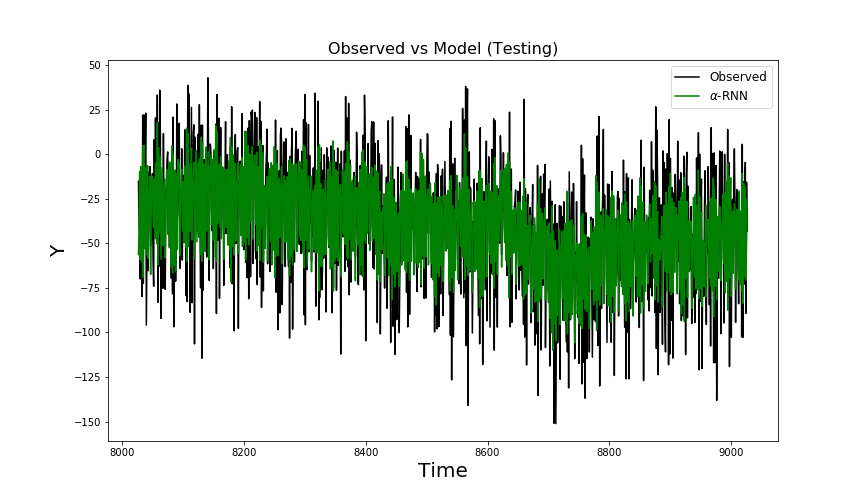} 
\caption{$\alpha$-RNN.}
\end{subfigure}
\begin{subfigure}[t]{0.31\columnwidth}
\centering
\includegraphics[width=\textwidth, height= 0.2\textheight]{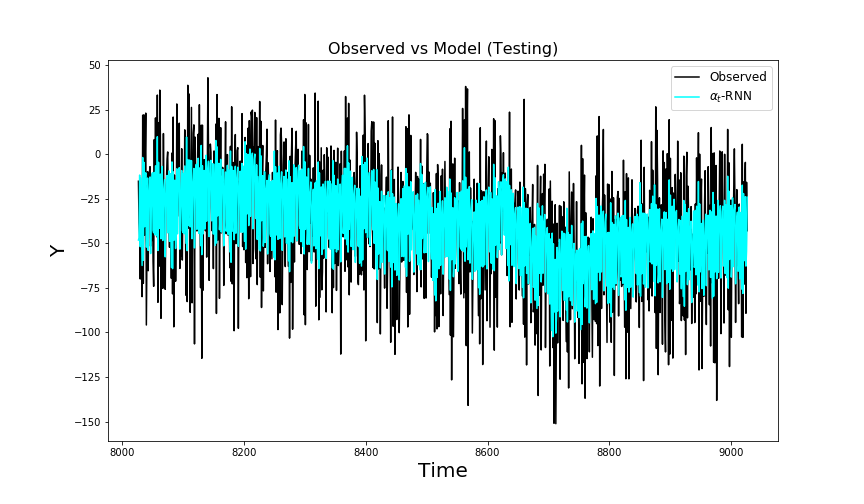} 
\caption{$\alpha_t$-RNN.}
\end{subfigure}
\begin{subfigure}[t]{0.31\columnwidth}
\centering
\includegraphics[width=\textwidth, height= 0.2\textheight]{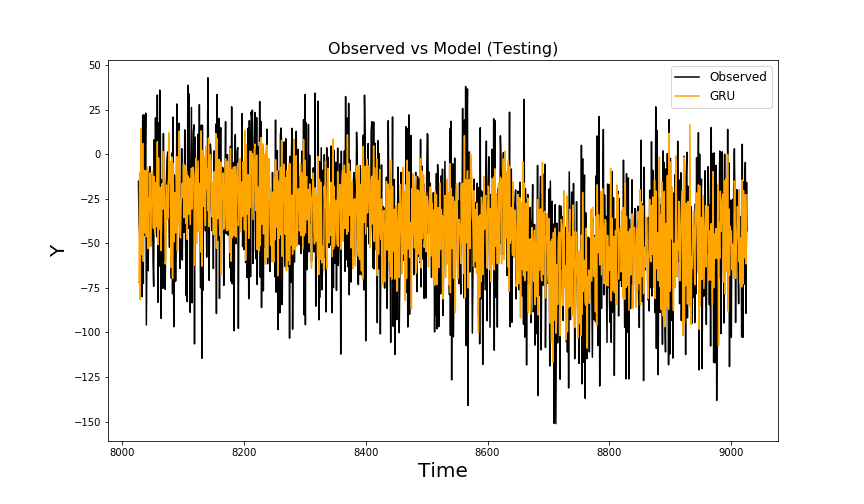} 
\caption{GRU.}
\end{subfigure}
\newline
\begin{subfigure}[t]{0.31\columnwidth}
\centering
\includegraphics[width=\textwidth, height= 0.2\textheight]{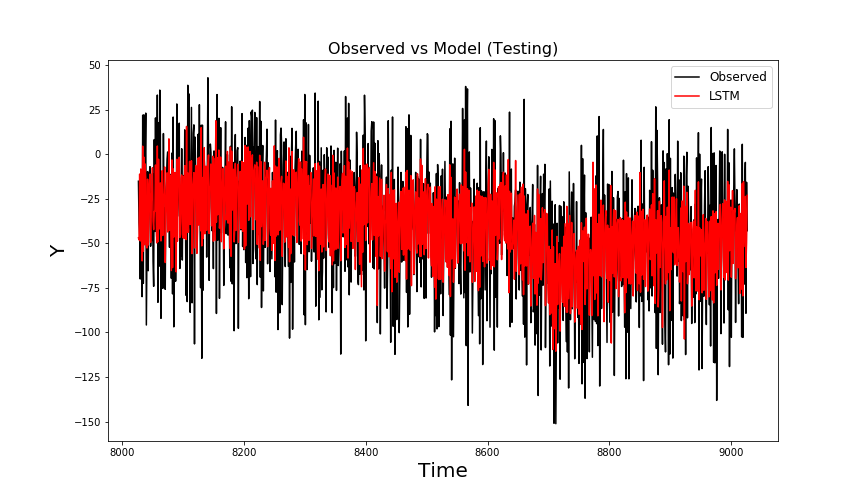} 
\caption{LSTM.}
\end{subfigure}
\begin{subfigure}[t]{0.6\columnwidth}
\centering
\includegraphics[width=0.8\textwidth, height= 0.2\textheight]{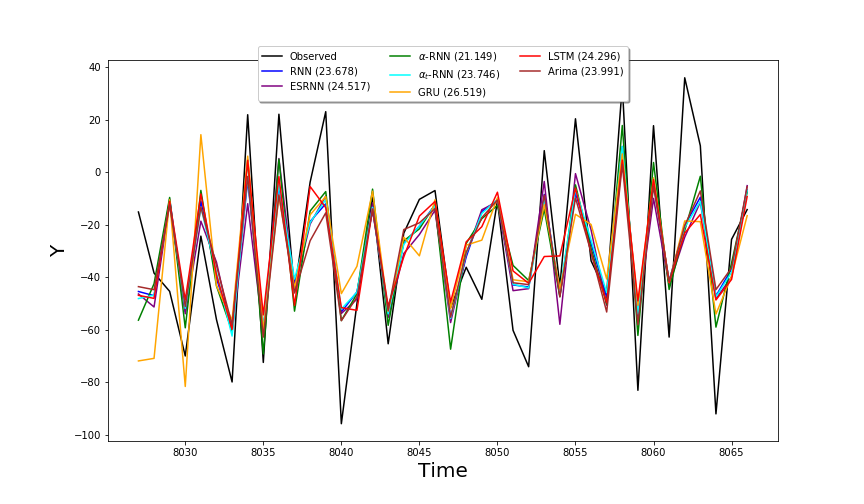} 
\caption{Comparison of 5-step ahead rolling forecasts.}
\end{subfigure}
\caption{\textit{The 5-step ahead rolling forecasts for various recurrent architectures and ARIMA applied to the out-of-sample LLM synthetic datset.}}
\label{fig:dcp}
\end{figure}

 We note that the $\alpha$-RNN exhibits the lowest RMSE by far and, consistent with the comment regarding the error plots, the GRU exhibits the highest RMSE. All methods except the GRU and LSTM outperform ARIMA and we note in passing that all methods consistently under-estimate the absolute value of the observations. Also, based on additional experiments provided on request, there is no improvement in the performance of the recurrent networks without seasonal and trend decomposition.

For the same experimental configuration - using five-step ahead rolling forecasts applied to seasonally and trend decomposed time series, Table \ref{tab:sea_results} compares the training and testing MSEs and MAEs in addition to training and prediction times.  The time series cross validated parameters $H$ and $\lambda_1$ are found using ten folds and the total number of trainable parameters are given. The optimal $\lambda_1$ is zero in each case and is hence not shown in the table. The $\alpha$-RNN has by far the lowest MSE and MAE on the test set, requires less parameters (memory) and training time is approximately 2-3x lower than GRUs and LSTMs. The GRU shows the strongest tendency to over-fit, the variance is the largest and the in-sample bias is the lowest. In fact, both the GRU and the LSTM are outperformed by ARIMA. ARIMA requires evaluation of the moving average sub-model, which is excessively time consuming for multi-step forecasting over longer test sets.

  \begin{table}[H]
	\centering
\resizebox{\columnwidth}{!}{%
\begin{tabular}{ccccccccc}
\hline
Architecture & Parameters & H & MSE (train) &  MSE (test) & MAE (train) & MAE (test) & training time (s) & prediction time (s)  \\
\hline
RNN&41&5&411.333&560.633&16.186&18.725&151.784&7.928\\
ESRNN&42&5&410.13&562.804&16.166&18.7651&409.246&8.512\\
$\alpha$-RNN&132&10&356.514&447.28&15.126&16.839&353.273&11.018\\
$\alpha_t$-RNN&86&5&409.655&563.885&16.166&18.835&894.411&12.31\\
GRU&1341&20&350.744&703.246&14.865&21.057&1135.56&15.625\\
LSTM&491&10&396.696&590.312&15.887&19.215&824.473&12.584\\
ARIMA(2,0,2)&5&0&-&575.572&-&18.957&0.076&1365.387\\

 \hline
  \end{tabular}
 }
  \caption{\textit{The cross-validated parameters and performance measurements of the \textbf{five-step} ahead rolling forecasts on seasonally and trend decomposed time series from the LLM synthetic dataset. The half-life of the $\alpha$-RNN is found to be 1.077 observation periods ($\hat{\alpha}=0.4744$).}}
\label{tab:sea_results}
  \end{table}

 \subsection{Electricity consumption}
  $N=30000$ observations of hourly power system loads are provided over the period from January 1st 2008 to June 20th 2011 for the DK2 bidding zone collected in the open power systems time series data \citep{dk2}. The consumption time series is chosen as it exhibits both short term cyclical patterns and longer-term trends. However, while these cyclical patterns correspond to peak and off-peak consumption periods, the transitions are gradual. Without attempting to deseasonalize the data nor detrend it, we apply the various architectures as in the previous experiment.
  
  We reject the Null hypothesis of the augmented Dickey-Fuller test at the 99\% confidence level in favor of the alternative hypothesis that the data is stationary (contains no unit roots). The test statistic is  $-10.991$ and the p-value is $0$ (the critical values are 1\%: -3.431, 5\%: -2.862, and 10\%: -2.567). 
  
  \begin{figure}[H]
      \centering
      \includegraphics[width=0.6\textwidth]{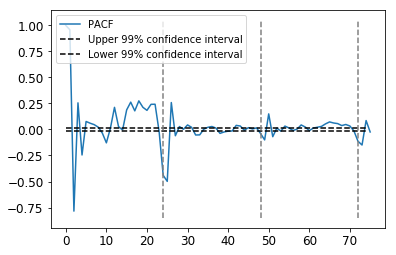}
      \caption{\textit{The PACF of the electricity load demand exhibits seasonality}}
      \label{fig:pacf_electricity}
  \end{figure}
  
  The PACF in Figure \ref{fig:pacf_electricity} is observed to exhibit seasonality, although the effect is less regular than in the LLM dataset of Section \ref{sect:results_llm}, due to the presence of multi-modal seasonality resulting from the difference in load demand on week day and weekend days. We choose a sequence length of $p=30$ and perform a ten-step ahead forecast to highlight the limitations of simple architectures. 
  
  Due to the presence of outliers in the training and testing sets, we choose the mean absolute error as the loss function. Such an error measure is more robust to outliers. We vary the number of hidden nodes in multiples of 5 between 5 and 100 during time series cross-validation with 5 folds. 
  
\begin{figure}[H]
 \caption{\textit{Comparison of 10-step ahead electricity load forecasts.}}
\begin{subfigure}[t]{0.31\columnwidth}
\centering
\includegraphics[width=\textwidth, height= 0.2\textheight]{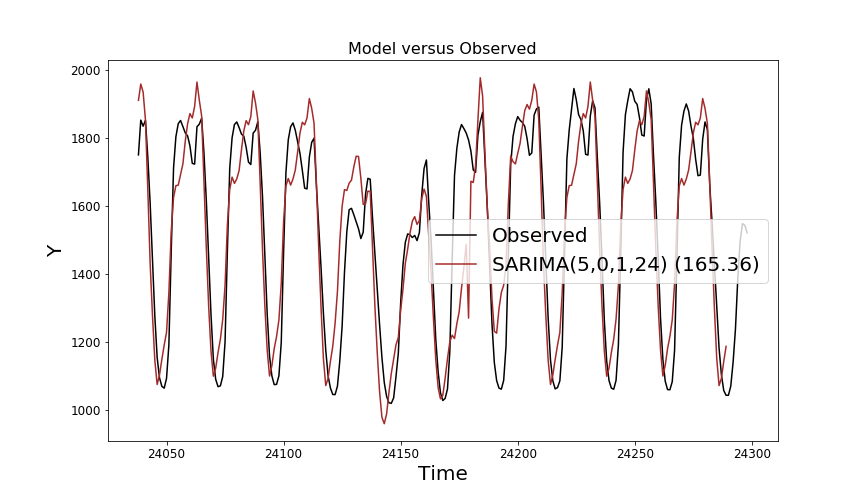} 
\caption{SARIMA.}
\end{subfigure}
\begin{subfigure}[t]{0.31\columnwidth}
\centering
\includegraphics[width=\textwidth, height=0.2\textheight]{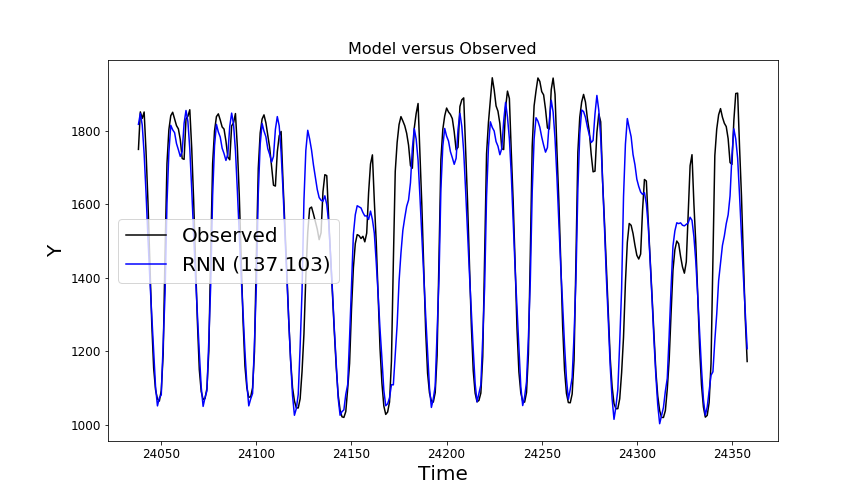} 
\caption{RNN.}
\end{subfigure}
\begin{subfigure}[t]{0.31\columnwidth}
\centering
\includegraphics[width=\textwidth, height= 0.2\textheight]{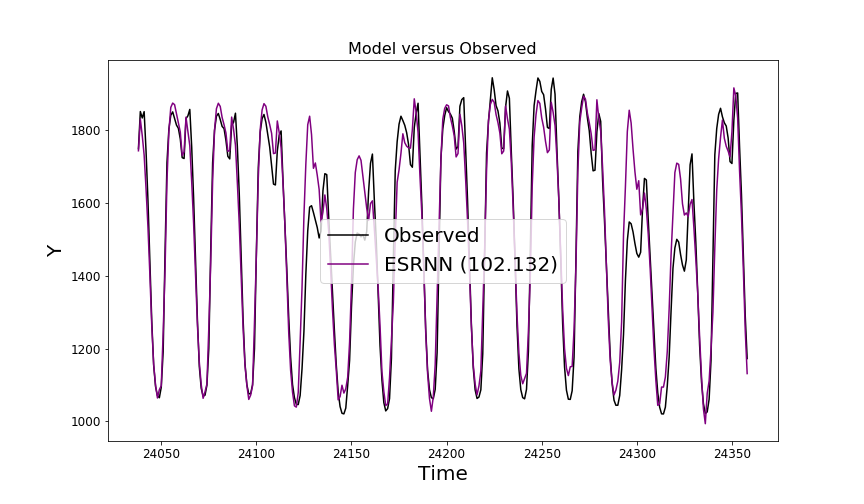} \caption{ES-RNN.}
\end{subfigure}
\newline
\begin{subfigure}[t]{0.31\columnwidth}
\centering
\includegraphics[width=\textwidth, height= 0.2\textheight]{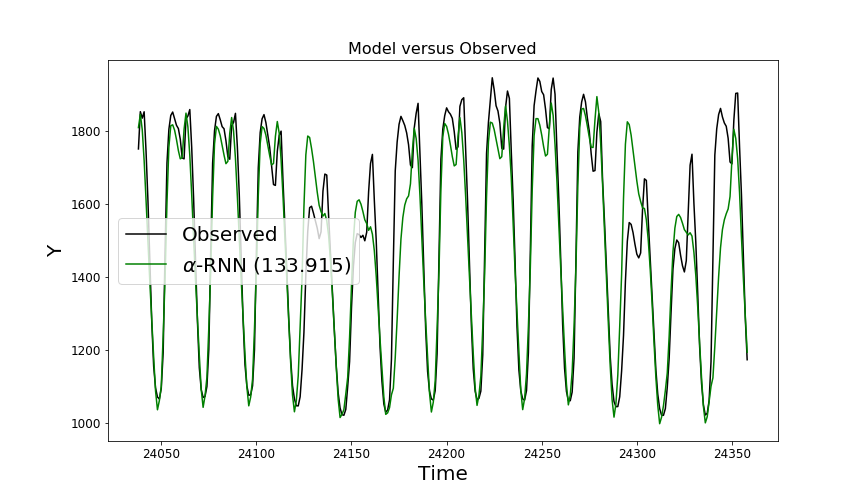} 
\caption{$\alpha$-RNN.}
\end{subfigure}
\begin{subfigure}[t]{0.31\columnwidth}
\centering
\includegraphics[width=\textwidth, height= 0.2\textheight]{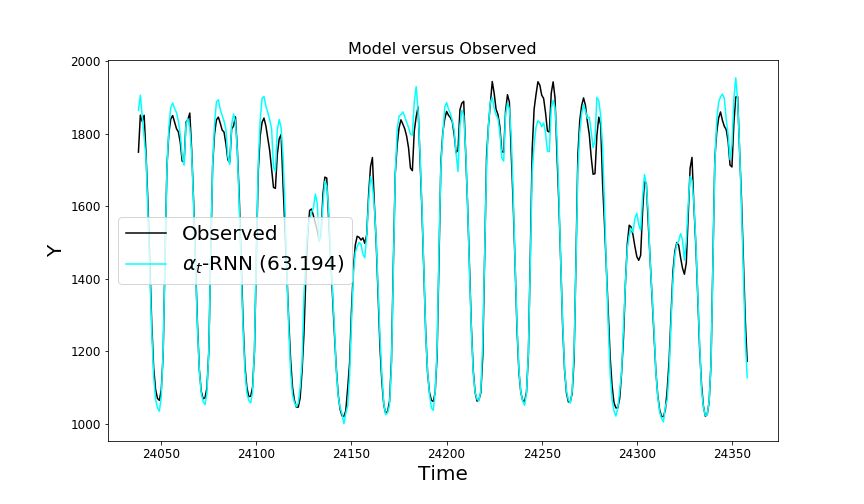} 
\caption{$\alpha_t$-RNN.}
\end{subfigure}
\begin{subfigure}[t]{0.31\columnwidth}
\centering
\includegraphics[width=\textwidth, height= 0.2\textheight]{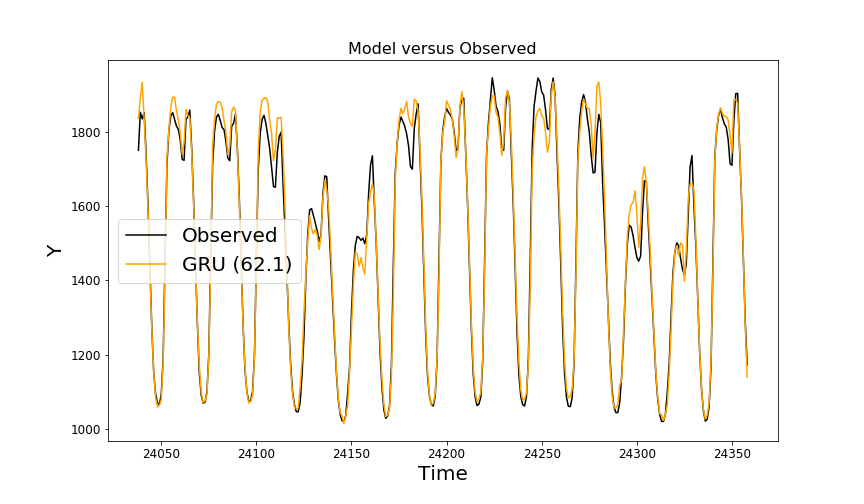} 
\caption{GRU.}
\end{subfigure}
\newline
\begin{subfigure}[t]{0.31\columnwidth}
\centering
\includegraphics[width=\textwidth, height= 0.2\textheight]{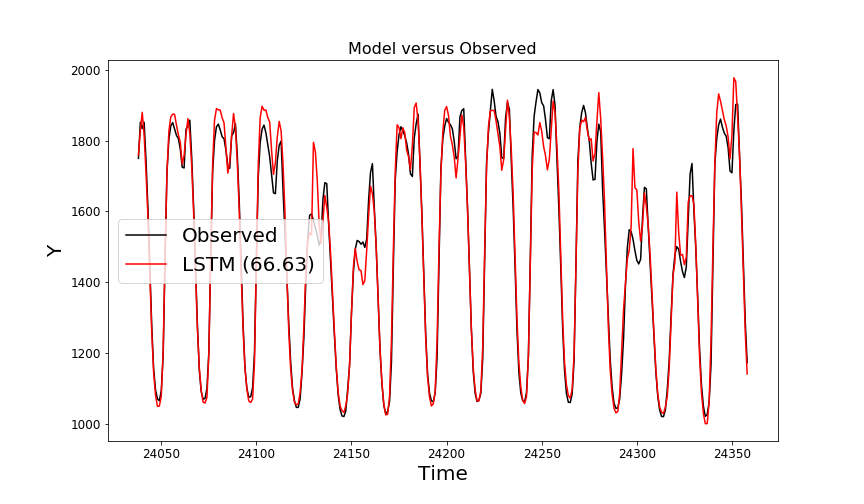} 
\caption{LSTM.}
\end{subfigure}
\begin{subfigure}[t]{0.69\columnwidth}
\centering
\includegraphics[width=0.7\textwidth, height= 0.2\textheight]{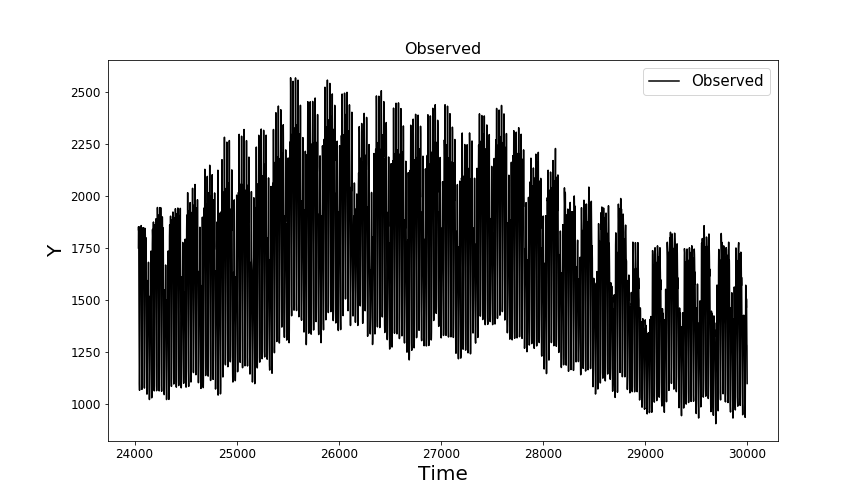} 
\caption{Observed test set.}
\end{subfigure}

\label{fig:exp2}
\end{figure}

  For comparison, we fit a SARIMA model and find the optimal model order to be $p=5, d=0, q=1$ with $s=24$. Due to severe memory and computational requirements during prediction, we use the 200 most recent observations in the training set for training.
  Figure \ref{fig:exp2}(a)-(g) compares the performance of the various networks over a shorter horizon of 360 hours. All methods are observed to capture the seasonal effects to varying degree, although the SARIMA followed by the RNN has the most difficulty in forecasting the shape of the load peaks, particularly on weekend days where the load profile peaks at lower values. The four stationary methods, SARIMA, RNNs, ES-RNNs and $\alpha$-RNNs aren't able to adapt to the weekend profiles, where as the non-stationary methods are able to.  Figure \ref{fig:exp2}(h) shows the observed test set over the full horizon. The load data is seen to exhibit longer-term seasonal variation in addition to short-term seasonal effects, highlighting the challenge of forecasting with trends and complex seasonal effects.
  
  From Table \ref{tab:errors_electicity2}, we further observe significant differences in the out-of-sample performance of the various networks. The SARIMA, RNN and ES-RNN exhibit the largest out-of-sample MSE and MAE whereas the $\alpha_t$-RNN and GRU exhibit the lowest MSE and the LSTM exhibits the lowest MAE. The MSE of the LSTM is substantially lower on the training set suggesting that the LSTM is over-fitting more than the other methods. Conversely, we observe that the other methods are slightly under-fitting under the MSE, but not the MAE. This behavior can be attributed to point-wise error outliers which are magnified under the MSE and appear more dominant in the training set than the test set.
  

 \begin{figure}[H]
 \caption{\textit{Comparison of loss profiles.}}
\begin{subfigure}[t]{0.31\columnwidth}
\centering
\includegraphics[width=\textwidth, height=0.2\textheight]{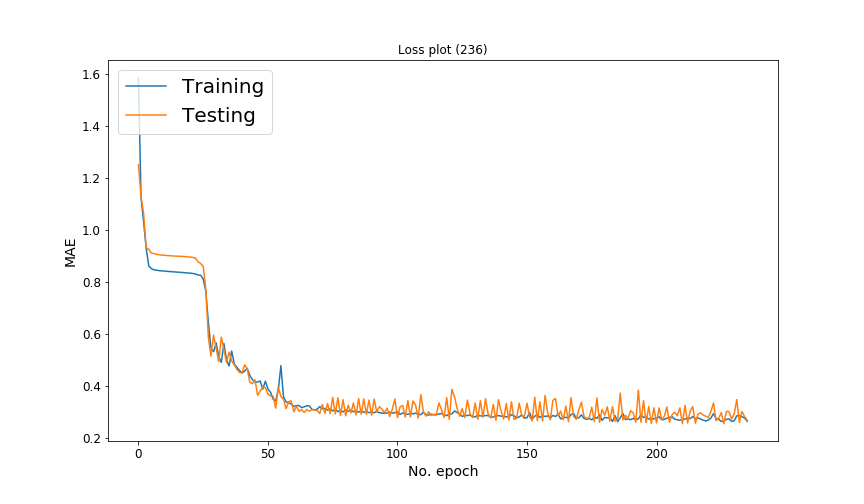} 
\caption{RNN.}
\end{subfigure}
\begin{subfigure}[t]{0.31\columnwidth}
\centering
\includegraphics[width=\textwidth, height= 0.2\textheight]{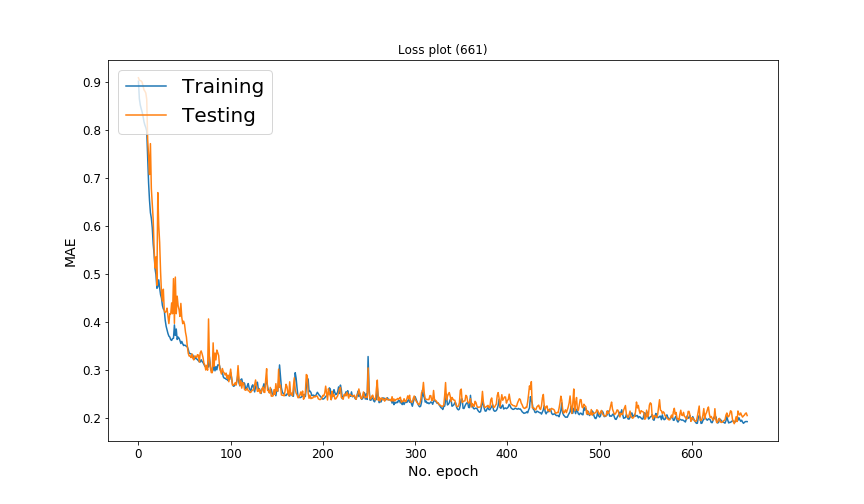} \caption{ES-RNN.}
\end{subfigure}
\begin{subfigure}[t]{0.31\columnwidth}
\centering
\includegraphics[width=\textwidth, height= 0.2\textheight]{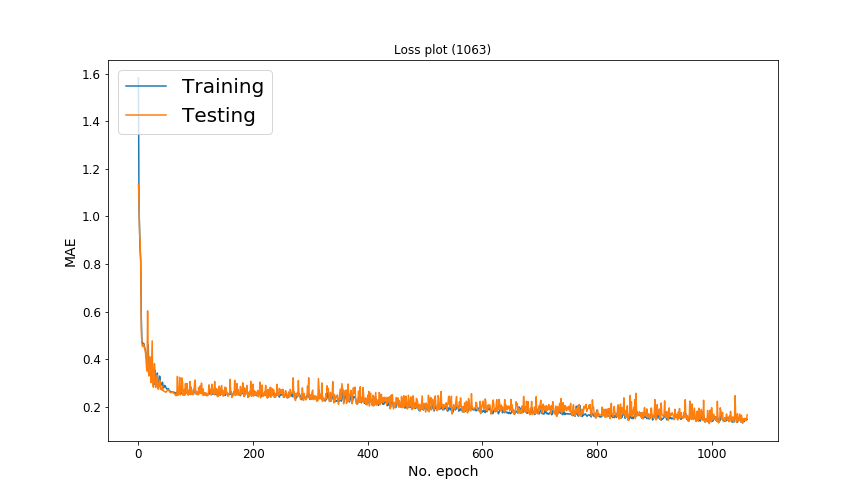} 
\caption{$\alpha$-RNN.}
\end{subfigure}
\newline
\begin{subfigure}[t]{0.31\columnwidth}
\centering
\includegraphics[width=\textwidth, height= 0.2\textheight]{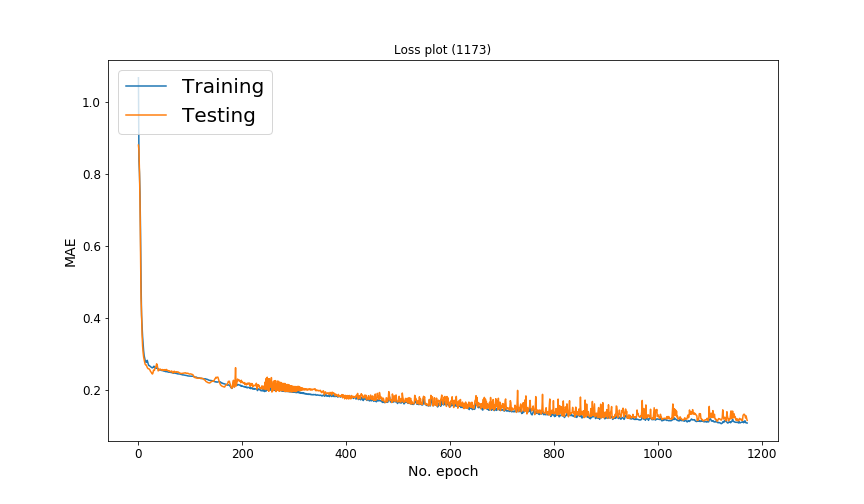}
\caption{$\alpha_t$-RNN.}
\end{subfigure}
\begin{subfigure}[t]{0.31\columnwidth}
\centering
\includegraphics[width=\textwidth, height=0.2\textheight]{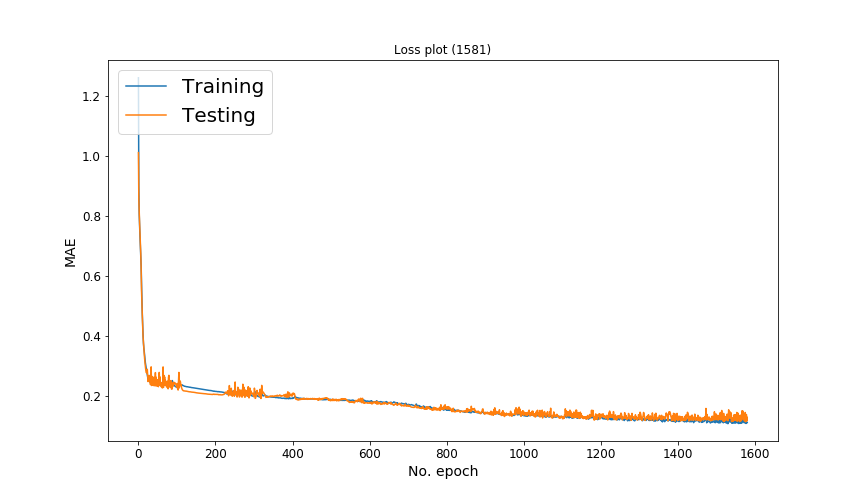} 
\caption{GRU.}
\end{subfigure}
\begin{subfigure}[t]{0.31\columnwidth}
\centering
\includegraphics[width=\textwidth, height= 0.2\textheight]{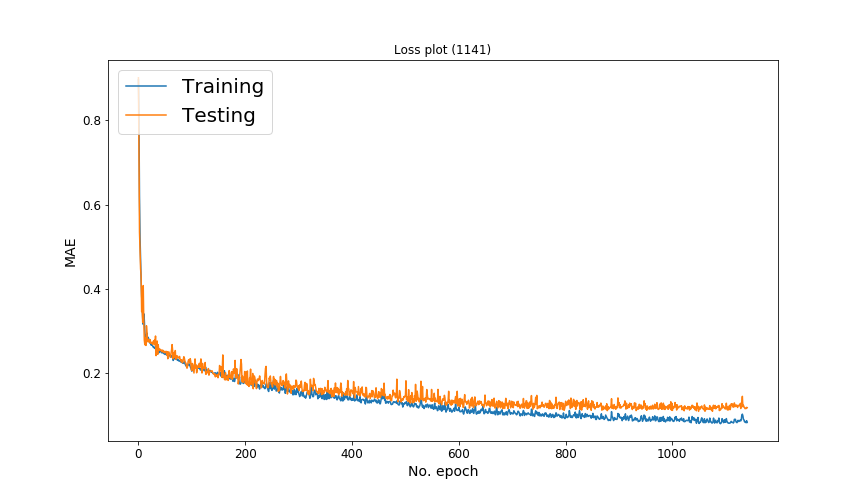} 
\caption{LSTM.}
\end{subfigure}
\label{fig:elec_convergence}
\end{figure}

 One advantage of the $\alpha$-RNN apparent from the training times is that it is able to solve the vanishing gradient problem without adding substantially increase training time. Even though the ES-RNN has the same number of trainable parameters as the the $\alpha$-RNN, the ES-RNN converges at a slower rate as seen in Figure \ref{fig:elec_convergence}. The number of epochs is shown in parenthesis above each figure. The testing error is observed to be more volatile than for the $\alpha$-RNN. Consequently the training time is almost 3x that of a plain RNN and almost 2x that of the $\alpha$-RNN. The GRU convergences at the slowest rate and combined with the large number of trainable parameters requires almost 40x the amount of time to train. The LSTM converges faster but also requires the most memory to store the parameters and exhibits some over-fitting (see Figure \ref{fig:elec_convergence}(f)). Hence, the $\alpha_t$-RNN presents itself as a lightweight dynamic network which trains more than 2x faster than a GRU and requires approximately half the number of parameters as a LSTM. The performance characteristics, however, are comparable.  

   \begin{table}[H]
	\centering
\resizebox{\columnwidth}{!}{%
\begin{tabular}{cccccccccc}
  \hline
  Architecture & Parameters & $\lambda_1$ & H & MSE (train) & MSE (test) & MAE (train) & MAE (test) & training time (s)\\
  \hline
  RNN&2651&0.001&50&19110.837&18797.331&92.175&94.438&331.727\\
ESRNN&2652&0.0&50&10033.155&10431.041&66.215&69.791&906.445\\
$\alpha$-RNN&10302&0.0&100&5707.978&5433.484&47.069&47.802&5139.467\\
$\alpha_t$-RNN&5351&0.0&50&4009.531&3993.478&38.331&42.152&5416.889\\
GRU&7851&0.0&50&3974.164&3856.419&38.896&42.03&11102.293\\
LSTM&40901&0.0&100&2381.012&4277.512&29.668&40.897&23652.981\\
SARIMA & 8 &- & - &-& 27343.803 & - & 125.869 & 41.383 \\
\hline
  \end{tabular}
  }
  \caption{\textit{The \textbf{ten-step} ahead electricity load forecasting models are compared for various architectures. The half-life of the $\alpha$-RNN is found to be 5.520 hours ($\hat{\alpha}=0.118$). }}
\label{tab:errors_electicity2}
  \end{table}


 \subsection{Short-term climate forecasting}
The Jena climate modeling dataset was recorded at the Weather Station operated by the Max Planck Institute for Biogeochemistry in Jena, Germany.  14 different quantities (such as air temperature, atmospheric pressure, humidity, wind direction etc) were recorded every 10 minutes, over several years. The dataset contains 420,551 observations covering the period 2009-2016 \citep{jena}. We demonstrate how the different networks forecast the temperature using all lagged observed variables. Each covariate in the training \emph{and} the test set is normalized using the moments of the training data only so as to avoid look-ahead bias or introduce a bias in the test data. 

\begin{figure}[H]
    \centering
    \includegraphics[width=0.8\textwidth]{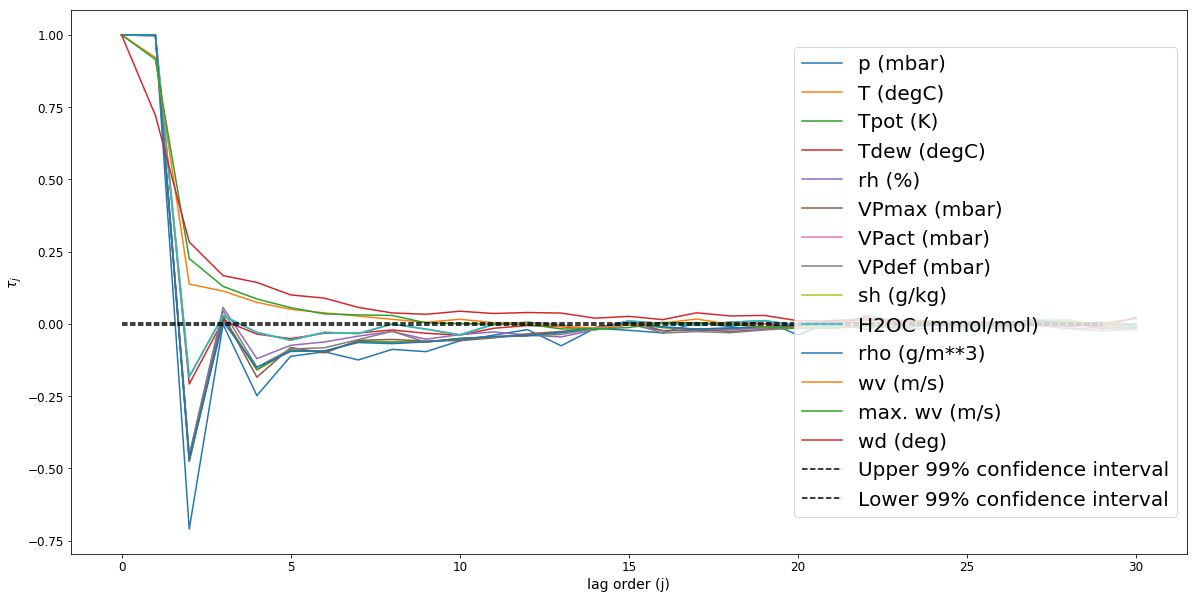}
    \caption{\textit{The partial autocorrelogram (PACF) for each of the covariates (features) used in the model. }}
    \label{fig:pacf_weather}
\end{figure}

We accept the Null hypothesis of Hansen's covariate-augmented Dickey-Fuller (CADF) test \citep{Hansen95} using the R package CADFtest \citep{CADFtest}. The test statistic is -19.392, the nuisance parameter $\rho^2=2.1068\times 10^{-5}$ and the p-value, $p=1$. The test is performed by regressing the temperature (Deg.C.) on to the other stationary covariates and their lags (up to a maximum lag of 20). Note that each covariate is tested individually first for stationarity (as required for the CADF test) using the augmented Dickey-Fuller test at the 99\% confidence level. The largest test statistic is  $-3.81841$ and the p-value is $0.002$ (the critical values are 1\%: -3.431, 5\%: -2.862, and 10\%: -2.567). We observe some evidence of cyclical memory in some of the covariates as seen in Figure \ref{fig:pacf_weather}, but these do not affect the stationarity. Thus, in summary, although each covariate is stationary, the resulting covariance between all the covariates is non-stationary.

\begin{figure}[H]
 \caption{\textit{Comparison of 10-step ahead temperature forecasts \& errors.}}
\begin{subfigure}[t]{0.31\columnwidth}
\centering
\includegraphics[width=\textwidth, height=0.2\textheight]{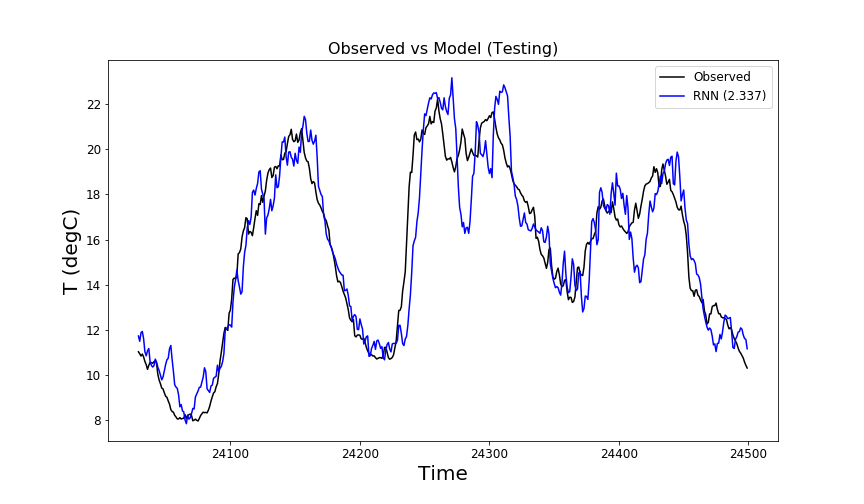} 
\caption{RNN.}
\end{subfigure}
\begin{subfigure}[t]{0.31\columnwidth}
\centering
\includegraphics[width=\textwidth, height= 0.2\textheight]{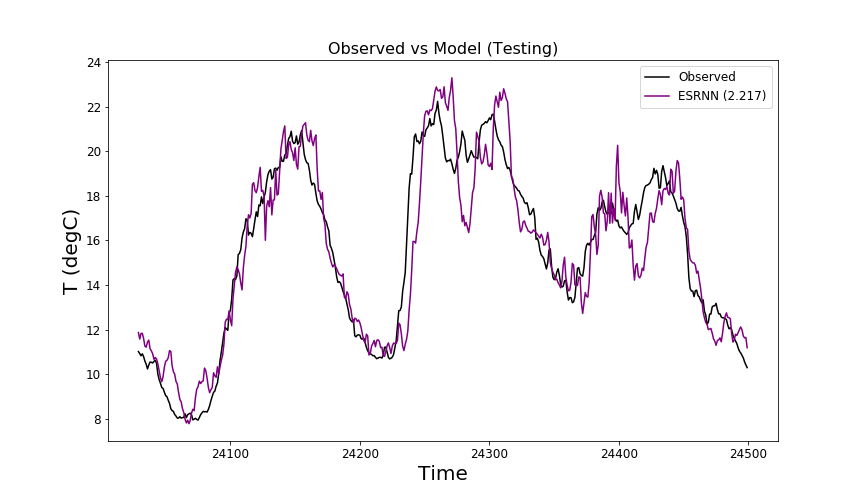} \caption{ES-RNN.}
\end{subfigure}
\begin{subfigure}[t]{0.31\columnwidth}
\centering
\includegraphics[width=\textwidth, height= 0.2\textheight]{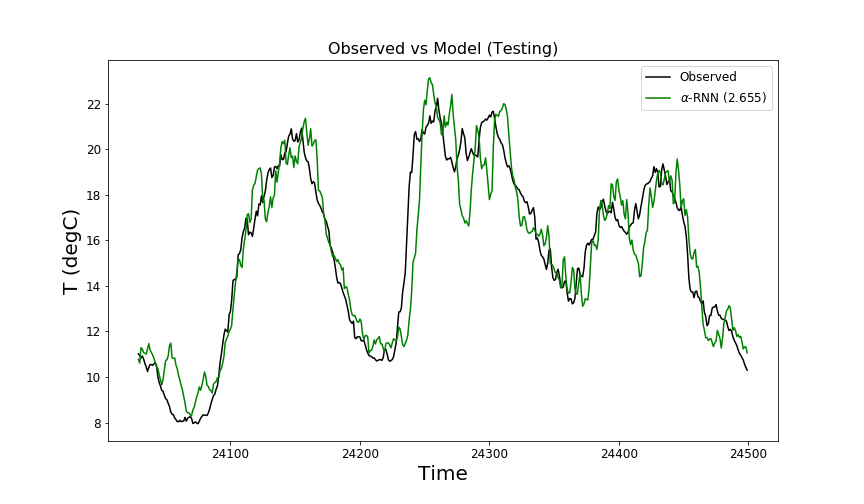} 
\caption{$\alpha$-RNN.}
\end{subfigure}
\newline
\begin{subfigure}[t]{0.31\columnwidth}
\centering
\includegraphics[width=\textwidth, height= 0.2\textheight]{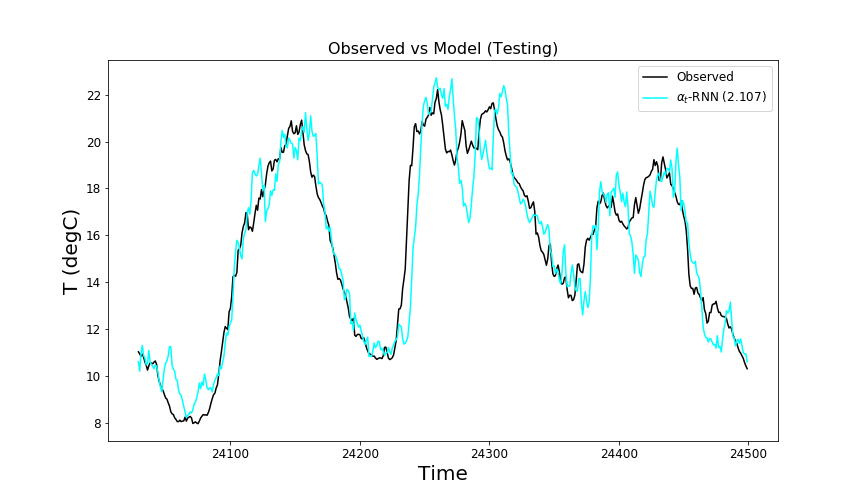} 
\caption{$\alpha_t$-RNN.}
\end{subfigure}
\begin{subfigure}[t]{0.31\columnwidth}
\centering
\includegraphics[width=\textwidth, height= 0.2\textheight]{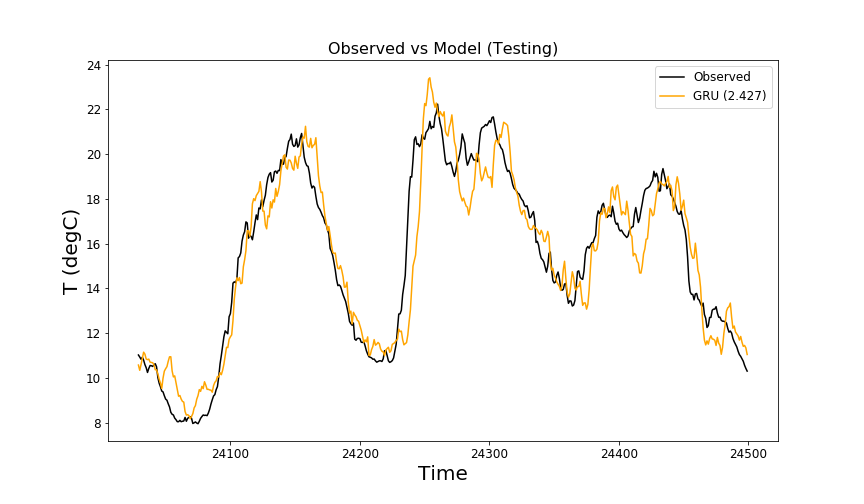} 
\caption{GRU.}
\end{subfigure}
\begin{subfigure}[t]{0.31\columnwidth}
\centering
\includegraphics[width=\textwidth, height= 0.2\textheight]{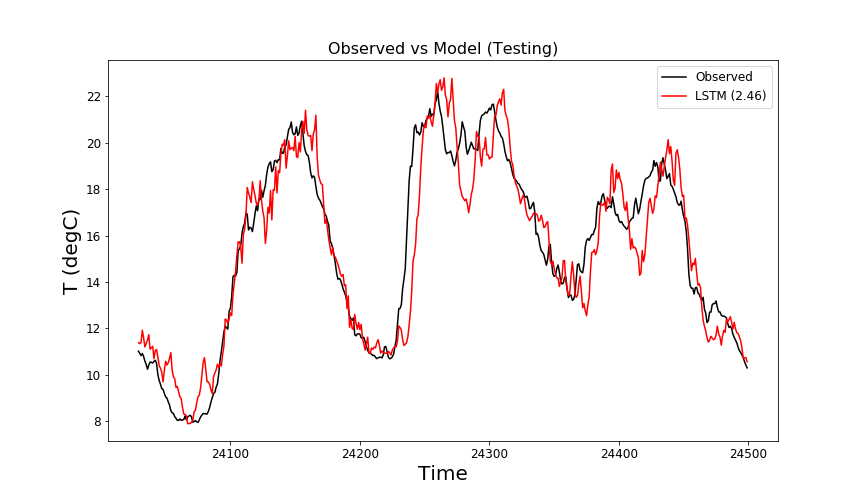} 
\caption{LSTM.}
\end{subfigure}
\newline
\begin{subfigure}[t]{0.48\columnwidth}
\centering
\includegraphics[width=0.8\textwidth, height= 0.25\textheight]{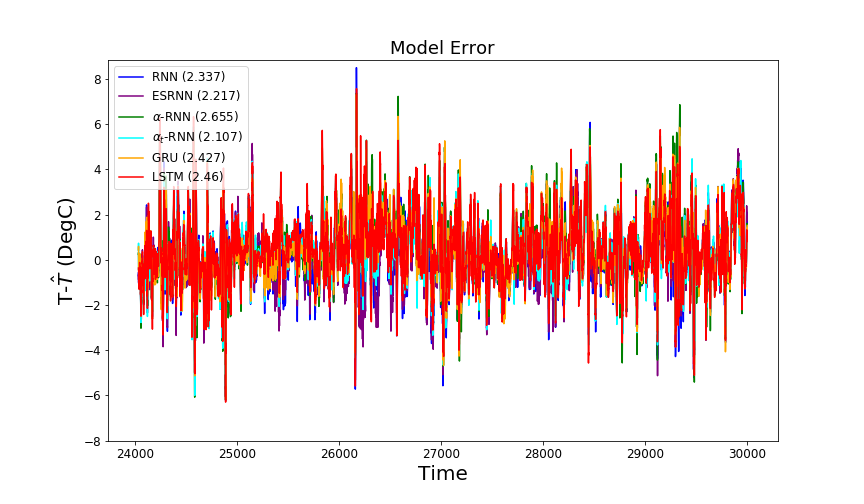} 
\caption{Error profiles.}
\end{subfigure}
\begin{subfigure}[t]{0.48\columnwidth}
\centering
\includegraphics[width=0.8\textwidth, height= 0.25\textheight]{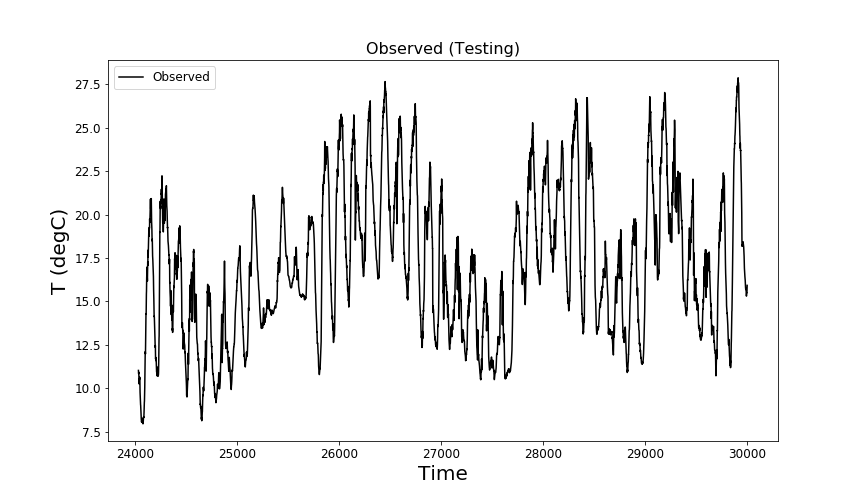} 
\caption{Observed test set.}
\end{subfigure}

\label{fig:weather}
\end{figure}

 We choose a sequence length of $p=20$ based on the PACF and perform a ten-step ahead forecast. Figure \ref{fig:weather} compares the 10 step ahead forecast error of the various recurrent neural networks and shows that the plain RNN exhibits the largest error in point estimates - there is an outlier shortly after observation 2000 in the test set. The other methods appear to exhibit similar error profiles except for the $\alpha_t$-RNN which exhibits the smallest error range. There is no apparent secular drift in any of the errors.  For completeness, Figure \ref{fig:weather}(g) shows the forecasts against the observed temperature (Deg.C.) over a 500 observation period.
 
Viewing the results of time series cross validation, using the first 23,971 observations, Table \ref{tab:errors_weather2} shows significant improvement of $\alpha_t$-RNN over the GRU and LSTM, suggesting that the reset gate and cellular memory provide no benefit for this forecasting problem. Rather the LSTM in particular overfits the training set, as does the $\alpha$-RNN in this case. The ES-RNN of \cite{SMYL202075}, performed without additional decomposition and using only a plain RNN, provides some improvement over a plain RNN, but much less so than the $\alpha_t$-RNN model.

 \begin{table}[H]
	\centering
\resizebox{\columnwidth}{!}{%
\begin{tabular}{cccccccc}
  \hline
  Architecture & Parameters & $\lambda_1$ & H & MSE (train) & MSE (test) & MAE (train) & MAE (test)\\
 \hline 
RNN&261&0&10&1.416 &2.337&0.891&1.151\\
ESRNN&263&0.00&10&1.382&2.252&0.876&1.129\\
$\alpha$-RNN&107&0.001&5&1.631&2.655&0.956&1.215\\
$\alpha_t$-RNN&216&0&5&1.409&2.107&0.882&1.097\\
GRU&761&0.001&10&1.644&2.427&0.971&1.178\\
LSTM&406&0&5&1.3625&2.46&0.845&1.168\\
\hline
  \end{tabular}
  }
  \caption{\textit{The \textbf{ten-step} ahead climate forecasts are compared for various architectures using time series cross-validation. The half-life of the $\alpha$-RNN is found to be 2.398 periods ($\hat{\alpha}=0.251$) or 23.98 minutes.}}
\label{tab:errors_weather2}
  \end{table}

\subsection{Stock price forecasting}
The purpose of this section is characterize the performance of five of the models in a Bayesian framework, thereby quantifying aleatoric uncertainty. Since the computational requirements for uncertainty quantification with RNNs are significantly larger than the frequentist setting, we shall restrict the study to a smaller dataset. Our dataset consists of $N=3020$ observations of daily adjusted close prices of IBM between 2006-01-03 and 2017-12-29.

\begin{figure}[H]
    \centering
    \includegraphics[width=0.6\textwidth]{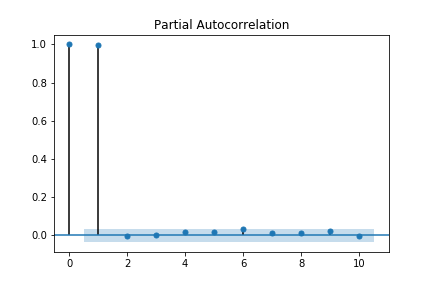}
    \caption{\textit{The partial autocorrelogram (PACF) for daily adjusted closing prices of IBM (USD) over the period 2006-01-03 to 2017-12-29. }}
    \label{fig:pacf_ibm}
\end{figure}

We demonstrate how the different networks forecast adjusted close prices up to 5 business days ahead using lagged observations of adjusted closing prices. The predictor in the training \emph{and} the test set is normalized using the moments of the training data only so as to avoid look-ahead bias or introduce a bias in the test data. 
We accept the Null hypothesis of the augmented Dickey-Fuller test as we can not reject it at even the 90\% confidence level. The data is therefore stationary (contains at least one unit root). The largest test statistic is  $-1.460$ and the p-value is $0.842$ (the critical values are 1\%: -3.962, 5\%: -3.412, and 10\%: -3.128). While the partial autocovariance structure is expected to be time dependent, we observe a short memory of only one lag by estimating the PACF over the entire history (see Figure \ref{fig:pacf_ibm}).

 We choose a sequence length of $p=2$ based on the PACF and perform a rolling five-step ahead forecast. This corresponds to the one-week ahead forecast, since there are 5 business days in a week. We comment in passing that a direct five-step ahead forecast is not advisable given the short sequence length, and hence we assess only the rolling forecast here, where we forecast and use intermediate forecasts over the five-day horizon. 
 
  The Bayesian RNNs are implemented using the Blitz PyTorch module. Cross-validation is performed over $\{5,10,15,20,25\}$ hidden neurons and $\lambda_1$ is varied over the set $\{10^{-3}, 10^{-2},10^{-1},0\}$. The optimal $\lambda_1$ is found to be zero for each model. The stopping criterion is different from the other experiments, reported in earlier sections. The training is stopped if there are 6 consecutive epochs where the loss (MSE) does not improve. The networks are fitted with an Adam optimizer with a learning rate of $0.001$. All initial network weights are treated as standard Gaussian iid random variables. We draw 10 samples for estimating ELBO and 10 samples from the fitted posteriors for prediction. 
 
 Figure \ref{fig:dcp} compares the forecasting performance of the various recurrent networks. In each plot the prediction is observed to drift from the observed series until the 5th day, after which the model input is reset with the latest data (rather than forecasted input). This provides insight in how quickly the forecast degrades with increasing prediction horizon. 
 
 The RNN, $\alpha$-RNN and GRU exhibit a saw-tooth like forecasted series which become more jagged the further the observation from the training period. The $\alpha_t$-RNN in Figure \ref{fig:dcp}(c) is much less jagged and remains so far out from the training set. Consequently it exhibits the lowest RMSE, as shown in the parenthesis of the legend. The 95\% confidence intervals are also shown and we observe that the observed series is frequently outside this interval in the case of the RNN and the GRU. Furthermore, the observed series falls increasingly outside of the confidence interval as points are further from the training set. The coverage is given in the parentheses of the interval legend. Conversely, the $\alpha_t$-RNN and $\alpha$-RNNs have higher coverages and the $\alpha_t$-RNN exhibits no trend of observations increasingly falling outside the confidence band.
 
 The comparison between the $\alpha$-RNN and $\alpha_t$-RNN is consistent with the non-stationary of the data- the $\alpha_t$-RNN is best able to capture the price dynamics by allowing the model's partial autocorrelation to be non-stationary.
 
 Table \ref{tab:errors_ibm} compares the performance of the Bayesian recurrent networks by one-day ahead and five-day ahead forecasts. The next day 95\% confidence intervals are back-tested over the test set and found to be comparable across the different networks - they consistently underestimate the empirical confidence intervals and more closely correspond to the 90\% empirical confidence intervals. The $\alpha$-RNN exhibits the most accurate confidence intervals - only 8\% of the observations fell outside the 95\% confidence intervals. The $\alpha$-RNN RMSE and MAE is substantially lower then the other methods while requiring less training time than the GRU and the LSTM. The prediction time of the $\alpha_t$-RNN is less than the GRU but, despite exhibiting fewer trainable parameters, is slower than the LSTM. The reason why the LSTM takes longer to train but is faster to predict is attributed to cellular memory caching implementation details beyond the scope of this paper.

\begin{figure}[H]\caption{\textit{Rolling price forecasts and confidence intervals for a prediction horizon of up to one week.}}
\begin{subfigure}[t]{0.49\columnwidth}
\centering
\includegraphics[width=0.8\textwidth, height=0.25\textheight]{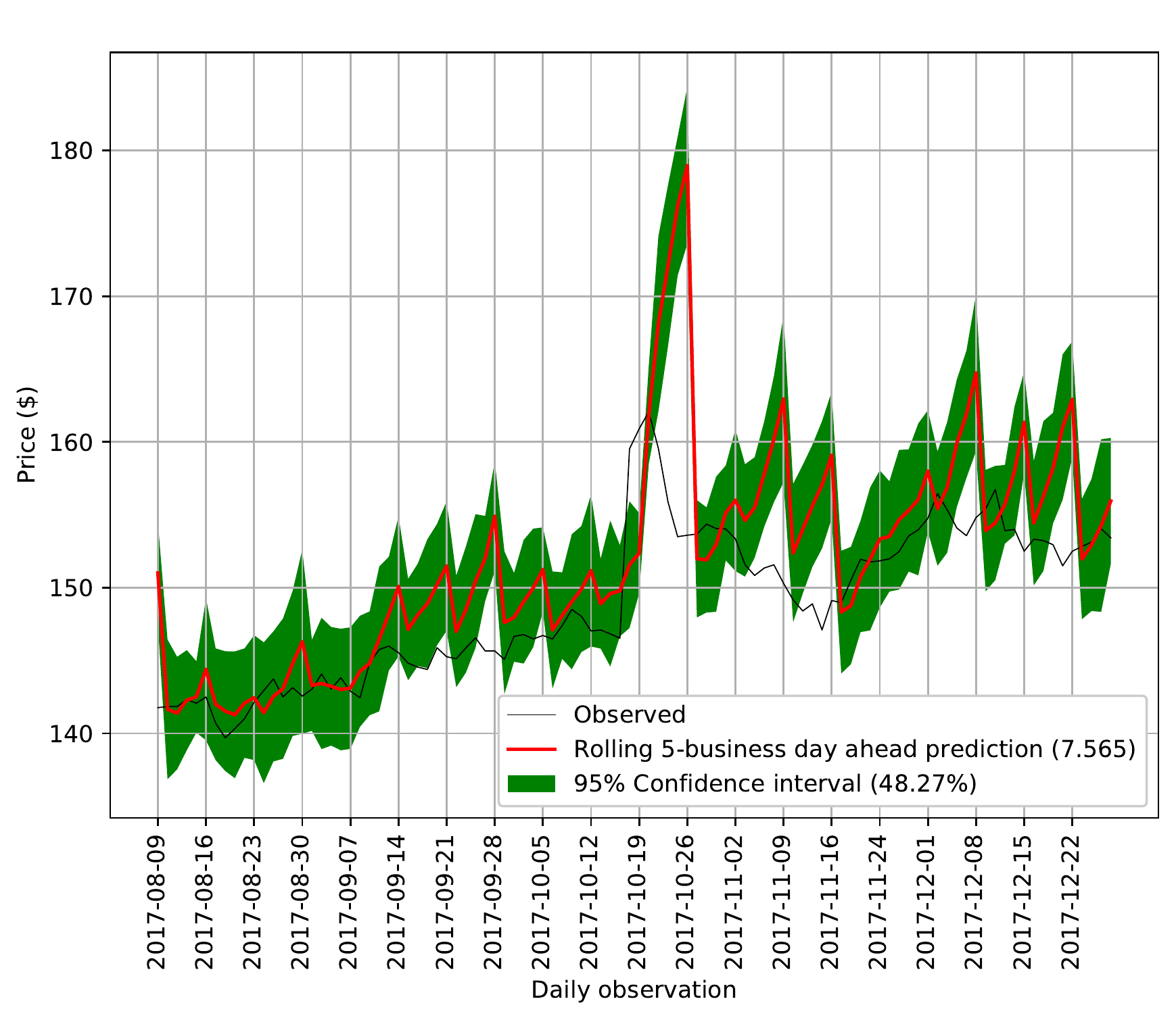} 
\caption{Bayesian RNN.}
\end{subfigure}
\begin{subfigure}[t]{0.49\columnwidth}
\centering
\includegraphics[width=0.8\textwidth, height= 0.25\textheight]{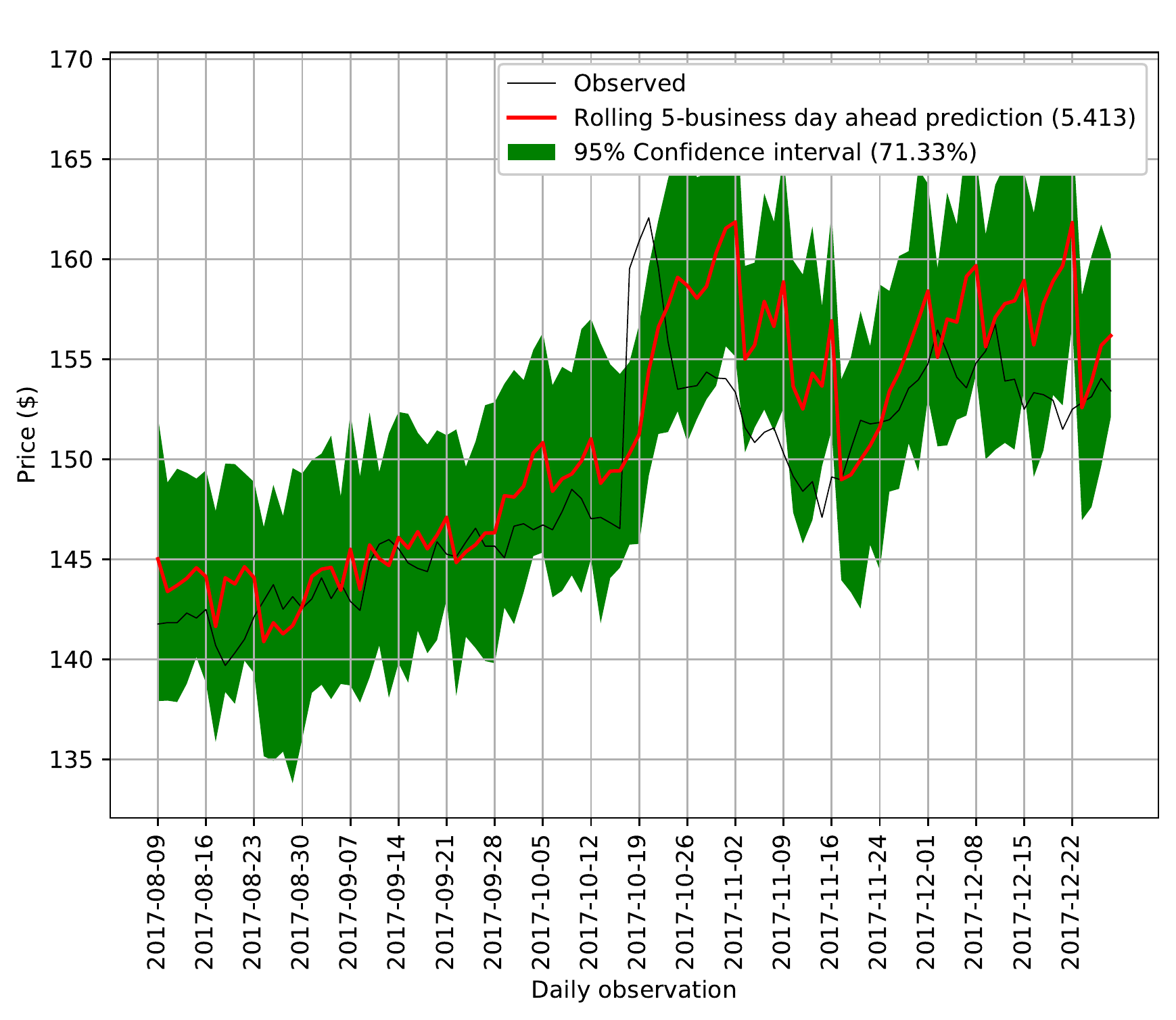} 
\caption{Bayesian $\alpha$-RNN.}
\end{subfigure}
\newline
\begin{subfigure}[t]{0.49\columnwidth}
\centering
\includegraphics[width=0.8\textwidth, height= 0.25\textheight]{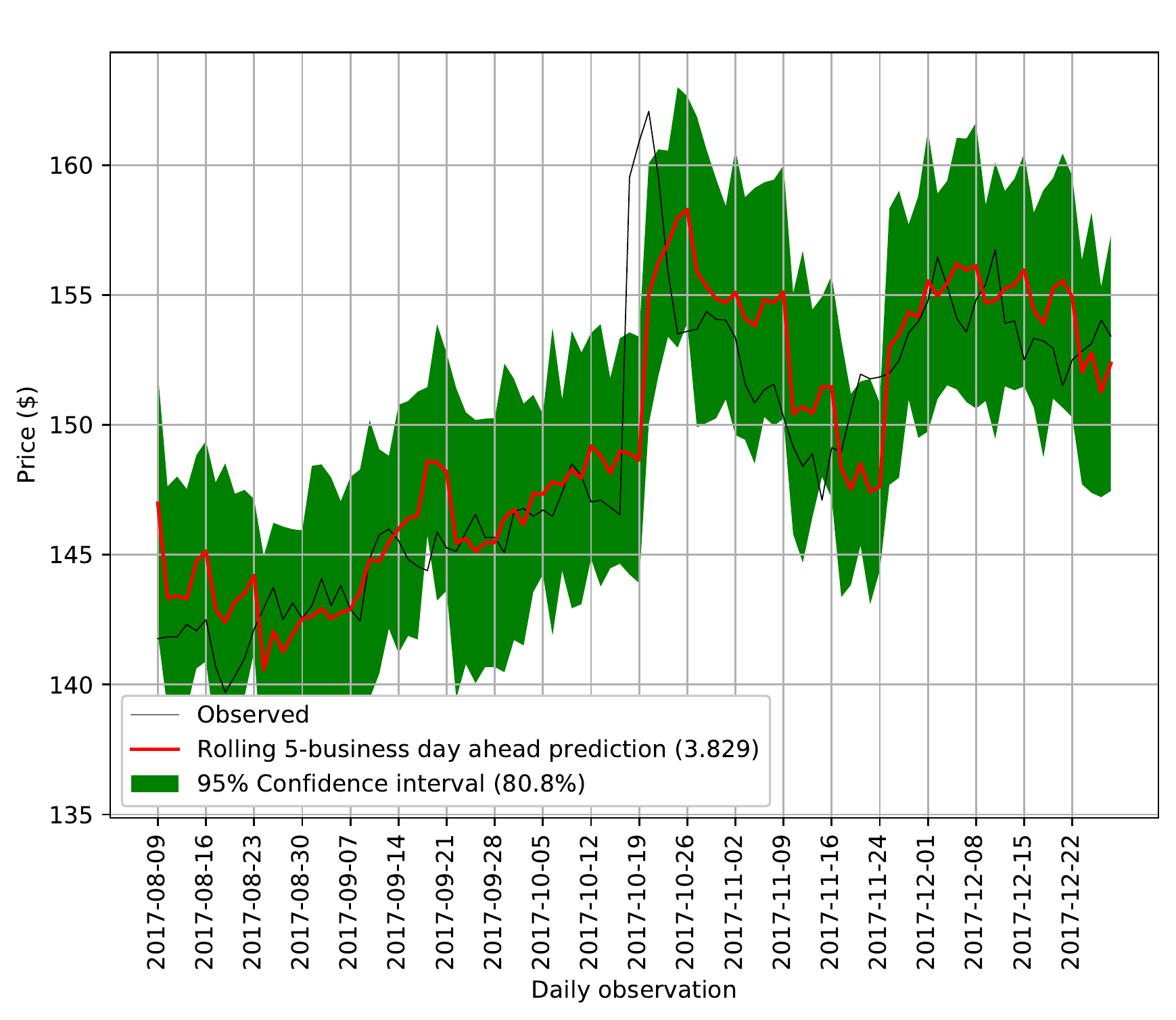} 
\caption{Bayesian $\alpha_t$-RNN.}
\end{subfigure}
\begin{subfigure}[t]{0.49\columnwidth}
\centering
\includegraphics[width=0.8\textwidth, height= 0.25\textheight]{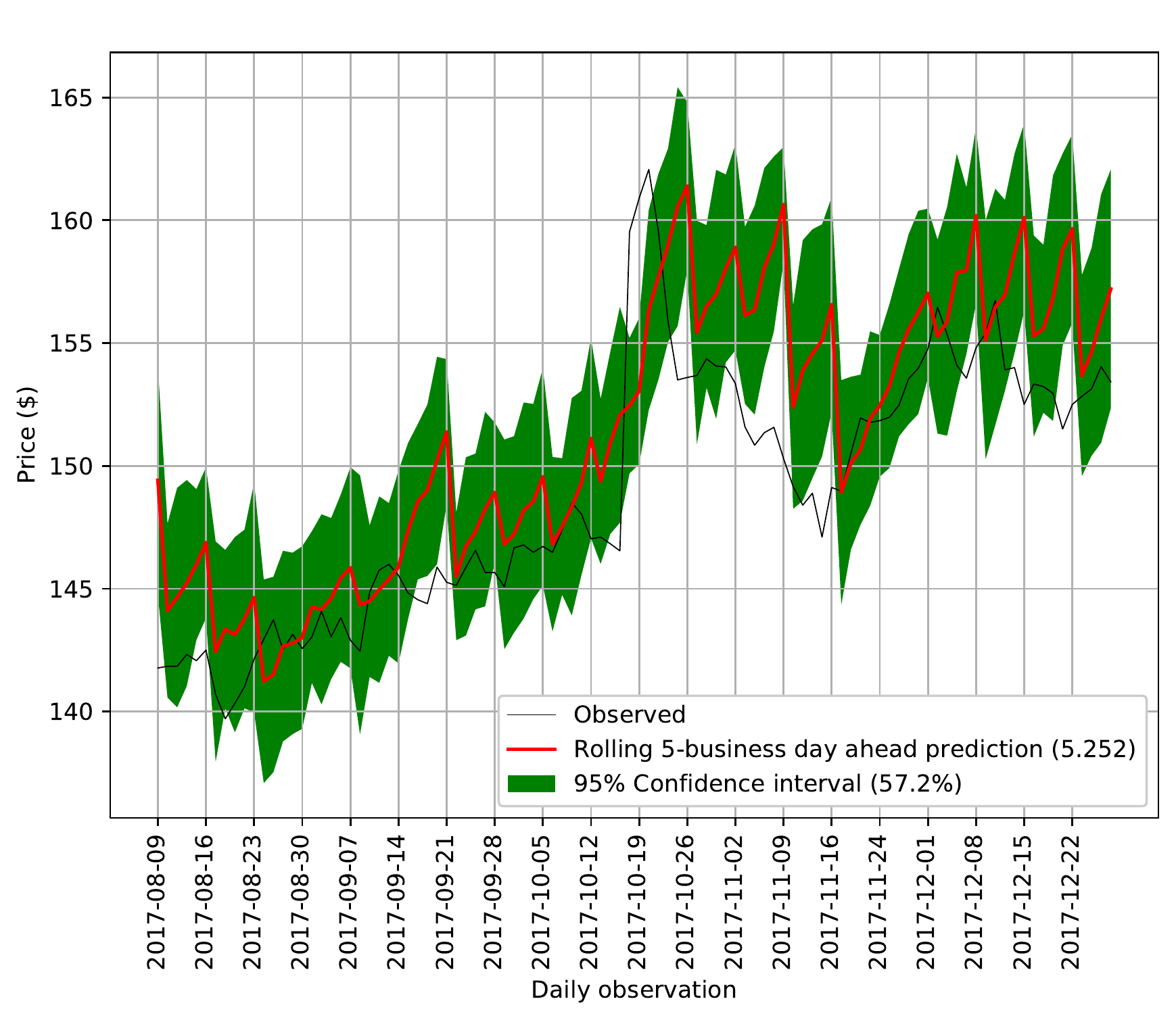} 
\caption{Bayesian GRU.}
\end{subfigure}
\newline

\begin{subfigure}[t]{0.49\columnwidth}
\centering
\includegraphics[width=0.8\textwidth, height= 0.25\textheight]{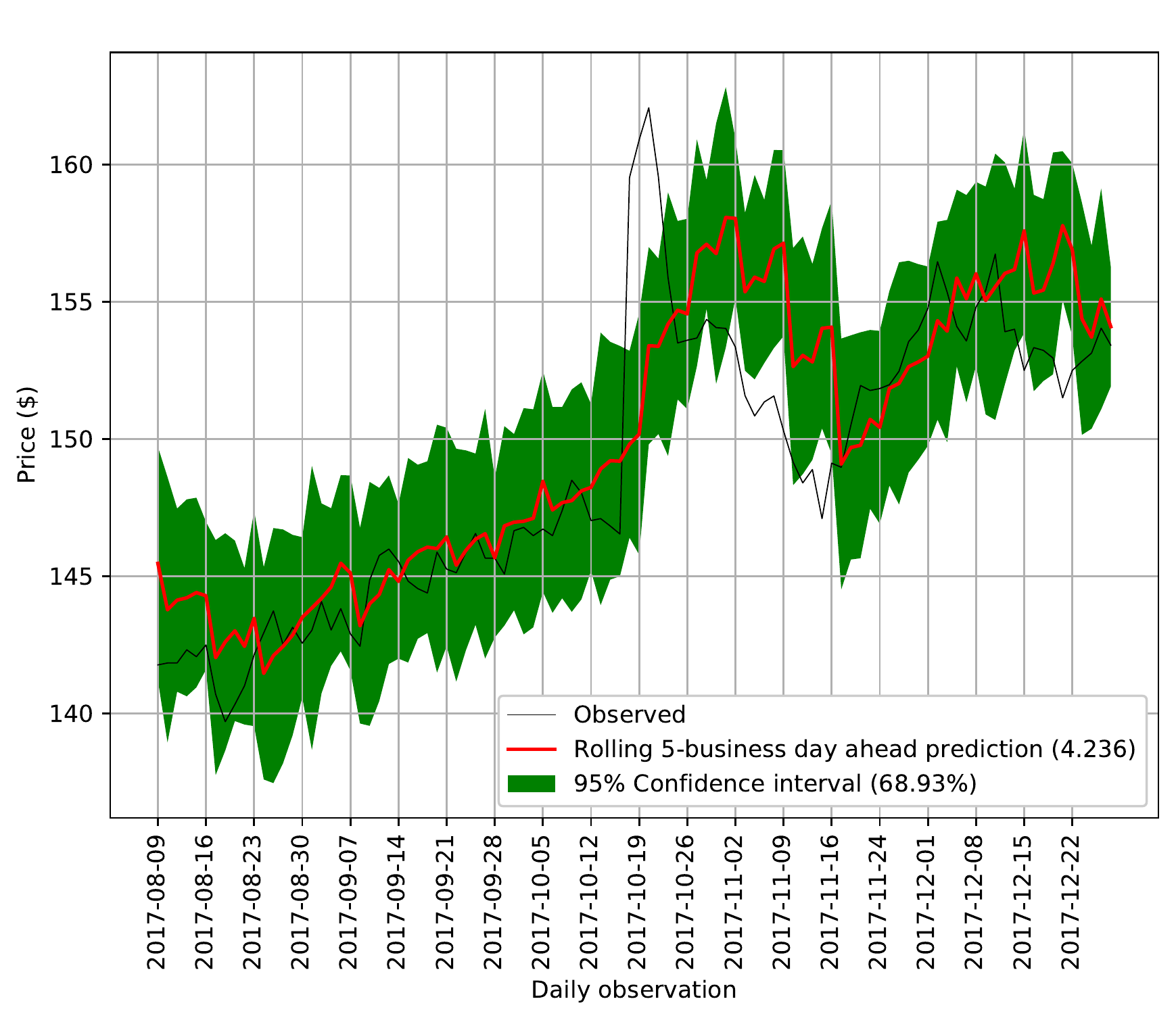} 
\caption{Bayesian LSTM.}
\end{subfigure}

\label{fig:dcp}
\end{figure}

 \begin{table}
	\centering
\resizebox{\columnwidth}{!}{%
\begin{tabular}{ccccccccc}
\hline
Architecture & Parameters & H & RMSE & MAE & Pred. Std. & Coverage & training time (s) & prediction time (s)  \\
\hline
One day ahead &&&&&&&&\\
\hline
RNN&126&10&3.093&2.271&147.91&0.879&64.842&55.716\\
$\alpha$-RNN&70&5&3.303&2.513&148.569&0.92&61.473&79.287\\
$\alpha_t$-RNN&166&10&2.743&1.968&147.738&0.895&114.779&99.688\\
GRU&306&15&3.038&2.278&147.791&0.868&129.999&112.149\\
LSTM&206&10&2.914&2.116&147.962&0.895&172.937&80.242\\
\hline
Five day ahead &&&&&&&&\\
\hline
RNN&126&10&7.565&5.848&147.946&0.483&64.842&55.5\\
$\alpha$-RNN&70&5&5.413&4.356&148.602&0.713&61.473&79.426\\
$\alpha_t$-RNN&166&10&3.829&2.835&148.098&0.808&114.779&95.868\\
GRU&306&15&5.252&4.206&147.796&0.572&129.999&104.856\\
LSTM&206&10&4.236&3.329&147.719&0.689&172.937&77.486\\
\hline
  \end{tabular}
  }
  \caption{\textit{The \textbf{one-step} and \textbf{five-step} ahead rolling forecasts are compared for various Bayesian recurrent networks. The half-life of the $\alpha$-RNN is found to be 0.508 days ($\hat{\alpha}=0.744$).}}
\label{tab:errors_ibm}
  \end{table}

  


  \section{Conclusion}\label{sect:conclusion}
  This paper presents a general class of exponential smoothed recurrent neural networks (RNNs) which are well suited to modeling non-stationary dynamical systems arising in industrial applications such as electricity load management and short-term weather prediction. In particular, when cast into a time series framework, we demonstrated how they characterize the non-linear partial autocorrelation structure of time series and directly capture dynamic effects such as seasonality and varying treads.  In addition to modeling complex data, these methods scale to large numbers of covariates and long time series. 
  
  We applied the Dickey-Fuller test or, in the case of multivariate data, the Hansen's test, to characterize the stationarity of the data. When the data is non-stationary we should expect, and indeed observe, superior performance of the dynamic architectures, the $\alpha_t$-RNN, GRU and LTSM generally show stronger performance on the Jena climate data and the stock forecasting data. When the data is stationary and void of any seasonal effects, such as the LLM synthetic dataset, we can expect and indeed observe modest gains or even inferior performance when using more complex networks. The $\alpha-$RNN model additionally offers the advantage of providing an interpretable smoothing parameter - it is shown to give the half-life of the model and hence the relative importance of more recent observations in the model.
  
  We additionally demonstrated how RNNs can provide uncertainty quantification through casting them in a Bayesian framework. Application of exponentially smoothed RNNs to electricity load demand,  climate, and stock prices highlighted the efficacy of exponential smoothing of the hidden-state for multi-step time series forecasting. In all of these examples, we show that $\alpha$-RNNs and $\alpha_t$-RNNs are well suited to forecasting, being much simpler than more complex architectures such as GRUs and LSTMs, yet retaining the most salient aspects needed for forecasting stationary and non-stationary series respectively. Because of their reduced size, they are faster to train and use for prediction, which is especially important in a Bayesian setting, where large number of simulations can easily result in computational challenges.
  
  The analysis presented in this paper considered only a subset of the many recent advancements in recurrent neural networks. Further research is required to analyze the effect of stacking networks on the partial auto-covariance and whether this permits more flexible decay profiles than the geometric decay seen in the linear RNN coefficients. Additionally, there are advances in the use of dilitation and dual attention mechanisms which respectively seek multi-scale and multivariate time series analysis to support the recent favorable empirical evidence presented using these techniques \citep{NIPS2017_6613}.

\bibliographystyle{chicago}
\bibliography{main}

\appendix

\section{Time Series Modeling Definitions}\label{sect:tsm}

\begin{defn}[Time series]\index{time series}
A time series is one observation of a stochastic process, over a specific
interval: $\{y_t\}_{t=1}^N$.
\end{defn}

\begin{defn}[Autocovariance] The $j^{th}$ autocovariance of a time series is 
$\gamma_{jt} := \mathbb{E}[(y_t-\mu_t)(y_{t-j}-
\mu_{t-j})]$ where $\mu_t:= \mathbb{E}[y_t]$.
\end{defn}

\begin{defn}[Covariance (weak) stationarity]
A time series is weak (or wide-sense) covariance stationary if it has
time constant mean and autocovariances of all orders:
\begin{eqnarray*}
\mu_t &=& \mu,\qquad~ \forall t\\
\gamma_{jt} &=& \gamma_j, \qquad~ \forall t.
\end{eqnarray*}
\end{defn}
As we've seen, this implies that $\gamma_j =\gamma_{-j}$: the autocovariances depend only on the interval between observations, but not the time of the observations.

\begin{defn}[Autocorrelation] The $j^{th}$ autocorrelation, $\tau_j$ is just the $j^{th}$ autocovariance divided by
the variance:
\be
\tau_j= \frac{\gamma_j}{\gamma_0}.
\ee
\end{defn}

\begin{defn}[White noise] White noise, $\phi_t$, is i.i.d. error which satisfies all three conditions:
\begin{itemize}
\item  $\mathbb{E}[\phi_t] = 0, \forall t$; 
\item  $\mathbb{V}[\phi_t] = \sigma^2, \forall t$; and 
\item  $\phi_t$ and $\phi_s$ are independent, $t \neq s, \forall t,s$. 
\end{itemize}
Gaussian white noise just adds a normality assumption to the error. White noise error is often referred to as a ``disturbance'', ``shock'' or ``innovation'' in the time series literature.
\end{defn}

\section{Proof of Theorem 1}\label{sect:PA}

Let's first consider a RNN(1) process, i.e. $\alpha=p=1$. The lag-1 partial autocovariance is
\be
\tilde\gamma_1=\mathbb{E}[y_t-\mu, y_{t-1}-\mu]=\mathbb{E}[\hat{y}_t + \epsilon_t- \mu, y_{t-1}-\mu],
\ee
and using the RNN(1) model with, for simplicity, a single recurrence weight, $\phi$:
\be
\hat{y}_{t}=\act(\phi y_{t-1})
\ee
gives
\be
\tilde\gamma_1=\mathbb{E}[\act(\phi y_{t-1}) + \epsilon_t- \mu, y_{t-1}-\mu]= \mathbb{E}[y_{t-1}\act(\phi y_{t-1})],
\ee
where we have assumed $\mu=0$ in the second part of the expression.


Continuing with the lag-2 autocovariance gives:  
\be\label{eq:12auto}
\tilde\gamma_2=\mathbb{E}[y_t-P(y_{t} \given y_{t-1}), y_{t-2}-P(y_{t-2} \given y_{t-1})],
\ee
and $P(y_{t} \given y_{t-1})$ is approximated by the RNN(1):

\be
\hat{y}_{t}=P(y_{t} \given y_{t-1})=\act(\phi y_{t-1}).
\ee
and $P(y_{t-2} \given y_{t-1})$ is approximated by the backward RNN(1):
\be
\hat{y}_{t-2}=P(y_{t-2} \given y_{t-1})=\act(\phi (\hat{y}_{t-1}+u_{t-1})),
\ee
so that we see, crucially, that $\hat{y}_{t-2}$ depends on $u_{t-1}$ but not on $\epsilon_t$. Substituting the backward RNN(1) and $u_{t}=y_{t}-\hat{y}_{t}$ into Equation \ref{eq:12auto} gives
\be
\tilde\gamma_2=\mathbb{E}[\epsilon_t, y_{t-2}-\act(\phi (\hat{y}_{t-1}+u_{t-1}))],
\ee
and $y_{t-2}-P(y_{t-2} \given y_{t-1})$ hence depends on $\{u_{t-1}, u_{t-2},\dots\}$. Thus we have that $\tilde\gamma_2=0$.

Now suppose $\alpha\in (0,1)$. Repeating the above we have the backward $\alpha$-RNN:
\be
\hat{y}_{t-2}=P(y_{t-2} \given y_{t-1})=\act(\phi( \alpha(\hat{y}_{t-1}+u_{t-1}) + (1-\alpha)\tilde{h}_t))
\ee
and we see that, by virtue of the dependency of $\tilde{h}_t$ on $y_t$ and hence $\hat{y}_t+\epsilon_t$, that the lag-2 autocovariance is no longer zero.

Now consider the lag-2 partial autocovariance of the RNN(2) process, again with $\alpha=1$. Using the backward RNN(2) model:
\be
\hat{y}_{t-2}=P(y_{t-2} \given y_{t-1})=\act\left(\phi\act(\phi (\hat{y}_{t} + \epsilon_t) + y_{t-1})\right),
\ee
which depends on $\epsilon_t$ and hence the lag-2 partial autocovariance:
\be
\tilde\gamma_2=\mathbb{E}[\epsilon_t, y_{t-2}-\act\left(\phi\act(\phi (\hat{y}_{t} + \epsilon_t) + \hat{y}_{t-1} + u_{t-1})\right)],
\ee
is not zero. It follows by induction that lag-s partial autocorrelations 
\be
\tilde\gamma_s=\mathbb{E}[\epsilon_t, y_{t-s}-P(y_{t-s} \given y_{t-s+1},\dots, y_{t-1})]=0, s>p,
\ee
since $P(y_{t-s} \given y_{t-s+1},\dots, y_{t-1})$ is approximated by the backward RNN(p):
\begin{eqnarray*}
\hat{y}_{t-s}&=&P(y_{t-s} \given y_{t-s+1},\dots, y_{t-1})\\
&=& \act\left(\phi\act(\phi\act(\dots, \phi\act(\hat{y}_{t-s+p} + u_{t-s+p}) + \dots \hat{y}_{t-s+p-1} + u_{t-s+p-1}) + \dots) + \hat{y}_{t-1} + u_{t-1}\right) 
\end{eqnarray*}
Thus the PACF for an $\alpha$-RNN(p) has a cut-off at $p$ lags when $\alpha=1$. With long memory, i.e. $\alpha\in(0,1)$, $\tilde{\tau}_s\neq 0, s>p$ and hence the minimum memory of $\alpha$-RNN(p) model with $\alpha\in(0,1]$ is p.




\section{Proof of Theorem 2}\label{sect:proof_stability}

The proof proceeds by induction. We first consider the RNN(1) model:
\be
y_t= \Phi^{-1}(L)[\epsilon_t]=\left(1-\act(\phi L)\right)^{-1} [\epsilon_t],
\ee
where for ease of exposition we have set $W_y=1, U_h=W_x=\phi\in\mathbb{R}$, and $b_h=b_y=0$ without loss of generality. Expressing this as a infinite dimensional non-linear moving average model
\be
y_t=\frac{1}{1-\act(\phi L)}[\epsilon_t]=\sum_{j=0}^{\infty} (\act(\phi L)[\epsilon_t])^j,
\ee
and the infinite sum will be stable when the $(\act(\cdot))^j$ terms do not grow with $j$, i.e. $|\act|< 1$ for all values of $\phi$ and $y_{t-1}$.  In particular, the choice $tanh$ satisfies the requirement on $\act$. For higher order models, we follow an induction argument and show first that for a RNN(2) model we obtain
\begin{eqnarray*}
y_t&=&\frac{1}{1-\act(\phi\act(\phi L^2) + \phi L)}[\epsilon_t]\\
&=&\sum_{j=0}^{\infty} \act^j(\phi\act(\phi L^2) + \phi L)[\epsilon_t],
\end{eqnarray*}
which again is stable if $|\act|< 1$ and it follows for any model order that the stability condition holds.

It follows that lagged unit impulses of the \emph{data} strictly decay
with the order of the lag when $|\act|< 1$. Again by induction, at lag 1, the output from the hidden layer is
\be
h_t=\act(\phi\mathbf{1} + \phi\mathbf{0})=\act(\phi\mathbf{1}).
\ee
The absolute value of each component of the hidden variable under a unit vector impulse at lag 1 is strictly less than 1: 
\be
|h_t|_j=|\act(\phi\mathbf{1})|_j < 1,
\ee
if $|\act(x)| <1$ and each element of $\phi\mathbf{1}$ is finite. Additionally if $\act$ is strictly monotone increasing then $|h_t|_j$ under a lag two unit innovation is strictly less than $|h_t|_j$ under a lag one unit innovation
\begin{eqnarray}
|\act(\phi\mathbf{1}))_j| &>& |\act(\phi\act(\phi\mathbf{1}))|_j.
\end{eqnarray}

The choice of tanh or sigmoid activation is therefore suitable for RNNs with finite weights and input.

\section{Relationship to GRUs and LSTMs} \label{sect:gru_lstm}
\subsection{GRUs} The $\alpha_t$-RNN model has no means to entirely reset its memory and become a feed-forward network (FFN). This is because the hidden variables update equation always depends on the previous smoothed hidden state, unless $U_{h}=0$. 
By adding a reset layer, we recover a GRU:

\begin{eqnarray*}
\text{smoothing}: \tilde{h}_t &=& \hat{\alpha}_t \circ\hat{h}_t + (1-\hat{\alpha}_t)\circ\tilde{h}_{t-1}\\
\text{smoother update}: \hat{\alpha}_t &=&\act^{(1)}(U_{\alpha}\tilde{h}_{t-1} + W_{\alpha}x_t+b_{\alpha})\\
\text{hidden state update}: \hat{h}_t &=&\act(U_{h}\hat{r}_{t}\circ\tilde{h}_{t-1} + W_{h}x_t+b_{h})\\
\text{reset update}: \hat{r}_t &=&\act^{(1)}(U_{r}\tilde{h}_{t-1} + W_{r}x_t+b_{r}).
\end{eqnarray*}
When viewed as an extension of our $\alpha_t$ RNN model, we observe that the effect of introducing a reset, or switch, $\hat{r}_t$, is to forget the dependence of $\hat{h}_t$ on the smoothed hidden state. Effectively, we turn the update for $\hat{h}_t$ from a plain RNN to a FFN and entirely neglect the recurrence. The recurrence in the update of $\hat{h}_t$ is thus dynamic. It may appear that the combination of a reset and adaptive smoothing is redundant. But remember that $\hat{\alpha}_t$ effects the level of error correction in the update of the smoothed hidden state, $\tilde{h}_t$, whereas $\hat{r}_t$ adjusts the level of recurrence in the unsmoothed hidden state $\hat{h}_t$.  Put differently, $\hat{\alpha}_t$ by itself can not disable the memory in the smoothed hidden state (internal memory), whereas $\hat{r}_t$ in combination with $\hat{\alpha}_t$ can. More precisely, when $\alpha_t=1$ and $\hat{r}_t=0$, $\tilde{h}_t=\hat{h}_t=\act(W_{h}x_t+b_{h})$ which is reset to the latest input, $x_t$, and the GRU is just a FFN. Also, when $\alpha_t=1$ and $\hat{r}_t>0$, a GRU acts like a plain RNN. Thus a GRU can be seen as a more general architecture which is capable of being a FFN or a plain RNN under certain parameter values. 

These additional layers (or cells) enable a GRU to learn extremely complex long-term temporal dynamics that a vanilla RNN is not capable of. Lastly, we comment in passing that in the GRU, as in a RNN, there is a final feedforward layer to transform the (smoothed) hidden state to a response:
\be
\hat{y}_t= W_Y\tilde{h}_t + b_Y.
\ee
\subsection{LSTMs}
The $\alpha_t$-RNN model, like the GRU, provides a mechanism for propagating a smoothed hidden state --- a long term memory which can be overridden and even turn the network into a plain RNN (with short memory) or even a memoryless FFN. More complex models using hidden units with
varying connections within the memory unit have been proposed in the engineering literature with empirical success \citep{Hochreiter1997, Gers2001,7926112}. LSTMs are similar to GRUs but have a separate (cell) memory, $c_t$, in addition to a hidden state $h_t$. LSTMs also do not require that the memory updates are a convex combination. Hence they are more general than exponential smoothing. The mathematical description of LSTMs is rarely given in an intuitive form, but the model can be found in, for example, \cite{Hochreiter1997}.

The cell memory is updated by the following expression involving a forget gate, $\hat{\alpha}_t$, an input gate $\hat{z}_t$ and a cell gate $\hat{c}_t$
\be\label{eq:c_update}
c_t = \hat{\alpha}_t \circ c_{t-1} + \hat{z}_t\circ \hat{c}_t.
\ee
In the terminology of LSTMs, the triple $(\hat{\alpha}_t, \hat{r}_t, \hat{z}_t)$ are respectively referred to as the forget gate, output gate, and input gate. Our change of terminology is deliberate and designed to provided more intuition and continuity with RNNs and the statistics literature. We note that in the special case when $\hat{z}_t=1-\hat{\alpha}_t$ we obtain a similar exponential smoothing expression to that used in our $\alpha_t$-RNN. Beyond that, the role of the input gate appears superfluous and difficult to reason with using time series analysis. 

When the forget gate, $\hat{\alpha}_t=0$, then the cell memory depends solely on the cell memory gate update $\hat{c}_t$. By the term $\hat{\alpha}_t \circ c_{t-1}$, the cell memory has long-term memory which is only forgotten beyond lag s if $\hat{\alpha}_{t-s}=0$. Thus the cell memory has an adaptive autoregressive structure.

The extra ``memory'', treated as a hidden state and separate from the cell memory, is nothing more than a Hadamard product: 
\be
h_t= \hat{r}_t\circ tanh(c_t),
\ee
which is reset if $\hat{r}_t=0$. If $\hat{r_t}=1$, then the cell memory directly determines the hidden state. 

Thus the reset gate can entirely override the effect of the cell memory's autoregressive structure, without erasing it. In contrast, the $\alpha_t$-RNN and the GRU has one memory, which serves as the hidden state, and it is directly affected by the reset gate.

The reset, forget, input and cell memory gates are updated by plain RNNs all depending on the hidden state $h_t$.
\begin{eqnarray*}
\text{Reset gate}: \hat{r}_t &=&\act(U_{r}h_{t-1} + W_{r}x_t+b_{r})\\
\text{Forget gate}: \hat{\alpha}_t &=&\act(U_{\alpha}h_{t-1} + W_{\alpha}x_t+b_{\alpha})\\
\text{Input gate}: \hat{z}_t &=&\act(U_{z} h_{t-1} + W_{z}x_t+b_{z})\\
\text{Cell memory gate}: \hat{c}_t &=&\tanh(U_{c} h_{t-1} + W_{c}x_t+b_{c}).
\end{eqnarray*}

Like the $\alpha_t$-RNN, the LSTM can function as a short-memory, plain-RNN; just set $\alpha_t=0$ in Equation \ref{eq:c_update}. However, the LSTM can also function as a coupling of FFNs; just set $\hat{r}_t=0$ so that $h_t=0$ and hence there is no recurrence structure in any of the gates. For avoidance of doubt, since the nomenclature doesn't suggest it, all models in this paper can model long and short-term autoregressive memory. The $\alpha_t$-RNN couples these memories through a smoothed hidden state variable. The LSTM separates out the long memory, stored in the cellular memory, but uses a copy of it, which may additionally be reset. Strictly speaking, the cellular memory has long-short autoregressive memory structure, so it would be misleading in the context of time series analysis to strictly discern the two memories as long and short (as the nomenclature suggests). The latter can be thought of as a truncated version of the former.

\section{Variational Inference} \label{sect:var}


Given the data $\mathcal{D} = (X,Y)$, the variational inference relies on approximating the posterior $p(\theta \mid \mathcal{D})$ with a more computationally tractable variation distribution  $q(\theta \mid \mathcal{D},\phi)$, where $\theta = (W,b)$. Then $q$ is found by minimizing the Kullback-Leibler divergence between the approximate distribution and the posterior, namely
\[
\text{KL}(q \mid\mid p) = \int q(\theta \mid \mathcal{D}, \phi)\log \dfrac{q(\theta \mid \mathcal{D}, \phi)}{p(\theta\mid \mathcal{D})}d\theta.
\]
Since $p(\theta\mid \mathcal{D})$ is not necessarily tractable, we replace minimization of $\text{KL}(q \mid\mid p) $ with maximization of the evidence lower bound (ELBO)\index{evidence lower bound}\index{ELBO|see {evidence lower bound}}
\[
\text{ELBO}(\phi) = \int q(\theta \mid \mathcal{D},\phi)\log \dfrac{p(Y\mid X,\theta)p(\theta)}{q(\theta \mid \mathcal{D}, \phi)}d\theta
\]
The $log$ of the total probability (evidence) is then
\[
\log p(D) =  \text{ELBO}(\phi) + \text{KL}(q \mid\mid p)\b{.}
\]
The sum does not depend on $\phi$, thus minimizing $\text{KL}(q \mid\mid p)$ is the same as maximizing $\text{ELBO}(q) $. Also, since $\text{KL}(q \mid\mid p) \ge 0$, which follows from Jensen's inequality, we have $\log p(\mathcal{D}) \ge  \text{ELBO}(\phi)$. Hence the reasoning behind the evidence lower bound nomenclature.  The resulting maximization problem $\text{ELBO}(\phi) \rightarrow \max_{\phi}$ is solved using stochastic gradient descent\index{stochastic gradient descent}.

To calculate the gradient, it is convenient to write the ELBO as
\[
\text{ELBO}(\phi) = \int q(\theta \mid \mathcal{D}, \phi)\log p(Y\mid X,\theta)d\theta - \int q(\theta \mid \mathcal{D},\phi) \log \dfrac{q(\theta\mid \mathcal{D}, \phi)}{p(\theta)}d\theta.
\]
The gradient of the first term $\nabla_{\phi}\int q(\theta \mid \mathcal{D}, \phi)\log p(Y\mid X,\theta)d\theta = \nabla_{\phi}E_q\log p(Y\mid X,\theta)$ is not an expectation and thus cannot be estimated using Monte Carlo methods. The key idea is therefore to represent the gradient $\nabla_{\phi} E_q\log p(Y\mid X,\theta)$ as an expectation of some random variable, so that Monte Carlo techniques can be used to calculate it. We use the following identity $\nabla_x f(x) = f(x) \nabla_x \log f(x)$ to obtain $\nabla_{\phi} E_q\log p(Y\mid \theta)$.
Thus, if we select $q(\theta \mid\phi)$ so that it is easy to compute its derivative w.r.t. $\phi$ and generate samples from it, the gradient can be efficiently calculated using Monte Carlo simulation.

\end{document}